\documentclass{article}
\usepackage[T1]{fontenc}

\usepackage{microtype}
\usepackage{graphicx}
\usepackage{subfigure}
\usepackage{booktabs} 

\usepackage{hyperref}

\usepackage{amsfonts,amssymb,amsmath,amsthm}

\newtheorem{theorem}{Theorem}
\newtheorem{assumption}{Assumption}
\newtheorem{lemma}{Lemma}
\newtheorem{corollary}{Corollary}

\newtheorem{definition}{Definition}

\newcommand{\squishlist}{
\begin{list}{{{\small{$\bullet$}}}}
{\setlength{\itemsep}{3pt}      \setlength{\parsep}{1pt}
\setlength{\topsep}{1pt}       \setlength{\partopsep}{0pt}
\setlength{\leftmargin}{1em} \setlength{\labelwidth}{1em}
\setlength{\labelsep}{0.5em} } }
\newcommand{\squishend}{  \end{list}  }

\usepackage[]{mathtools} 
\def\##1\#{\begin{align}#1\end{align}}
\def\$#1\${\begin{align*}#1\end{align*}}
\usepackage{enumitem}
\usepackage[]{amsthm} 
\usepackage[]{amssymb}
\usepackage{dsfont}
\usepackage{hyperref}
\usepackage{url}
\usepackage{xcolor}
\usepackage{amsmath,amsbsy}

\usepackage{tcolorbox}
\definecolor{grey}{rgb}{0.33, 0.33, 0.33}

\newcommand*\Bell{\ensuremath{\boldsymbol\ell}}

\newcommand{\p}{\mathbb{P}}
\newcommand{\bP}{\mathbb{P}}
\newcommand{\calD}{\mathcal D}

\newcommand{\bQ}{\mathbb{Q}}

\newcommand{\E}{\mathbb E}

\newcommand{\A}{\mathcal{A}}
\newcommand{\F}{\mathcal{F}}
\newcommand{\Pd}{\mathcal{P}}

\newcommand{\EE}{\mathcal{E}}
\newcommand{\R}{\mathbb{R}}

\newcommand{\err}{\mathsf{err}}
\newcommand{\D}{\mathcal D}

\newcommand{\ny}{\mathbf{\tilde{y}}}
\newcommand{\y}{\mathbf{y}}
\newcommand{\sy}{\mathbf{y}_{\text{LS}}}
\newcommand{\smy}{\mathbf{y}_{\text{LC}}}
\newcommand{\1}{\mathbf{1}}

\usepackage[accepted]{icml2021}

\icmltitlerunning{Understanding Instance-Level Label Noise: Disparate Impacts and Treatments}

\begin{document}

\twocolumn[
\icmltitle{Understanding Instance-Level Label Noise: \\ Disparate Impacts and Treatments}




\begin{icmlauthorlist}
\icmlauthor{Yang Liu}{UCSC}
\end{icmlauthorlist}

\icmlaffiliation{UCSC}{Department of Computer Science and Engineering, University of California, Santa Cruz, CA, USA}

\icmlcorrespondingauthor{Yang Liu}{yangliu@ucsc.edu}
\icmlkeywords{Disparate impact, learning with noisy labels, instance-level analysis}
\vskip 0.3in
]



\printAffiliationsAndNotice{}  

\begin{abstract}
This paper aims to provide understandings for the effect of an over-parameterized model, e.g. a deep neural network, memorizing instance-dependent noisy labels. We first quantify the harms caused by memorizing noisy instances, and show the \emph{disparate impacts} of noisy labels for sample instances with different representation frequencies.  We then analyze how several popular solutions for learning with noisy labels mitigate this harm at the instance level. Our analysis reveals that existing approaches lead to \emph{disparate treatments} when handling noisy instances. While higher-frequency instances often enjoy a high probability of an improvement by applying these solutions, lower-frequency instances do not. Our analysis reveals new understandings for when these approaches work, and provides theoretical justifications for previously reported empirical observations. This observation requires us to rethink the distribution of label noise across instances and calls for different treatments for instances in different regimes. 
\end{abstract}

\section{Introduction}

A salient feature of an over-parameterized model, e.g. a deep neural network, is its ability to memorize examples \cite{zhang2016understanding,neyshabur2017exploring}, and the memorization has proven to benefit the generalization performance \cite{arpit2017closer,feldman2020does,feldman2020neural}. Nonetheless, the potential existence of label noise, combined with the memorization effect, might lead to detrimental consequence \cite{song2019prestopping,yao2020searching,cheng2017learning,pmlr-v97-chen19g,han2020sigua,song2019prestopping}. In light of the reported empirical evidence of 
 harms caused by over-memorizing noisy labels,  we set out to understand this effect theoretically. 
Built on a recent analytical framework \cite{feldman2020does}, we demonstrate the varying effects of memorizing noisy labels associated with instances that sit at the different spectra of the instance distribution. 

Soon since the above negative effect was empirically shown, learning with noisy labels has been recognized as a challenging and important task. The literature has observed growing interests in proposing defenses, see \citet{natarajan2013learning,liu2016classification,menon2015learning,liu2019peer,lukasik2020does} and many more. The second contribution of this paper is to build an analytical framework to gain new understandings of how the existing solutions fare (Section \ref{sec:analysis}). While most existing theoretical results focus on the setting where label noise is homogeneous across training examples and focus on the distribution-level analysis, ours invests on the instance-level and aims to quantify when these existing approaches work and when they fail for different regimes of instances. Our result points out that while noisy labels for highly frequent instances contribute more to the drop of generalization power, they are also easier cases to fix with. We further highlight the need for taking additional care of long-tail examples \cite{zhu2014capturing}, where we prove existing solutions can have a substantial probability of failing. Our results call for immediate attention to a hybrid treatment of noisy instances.

To facilitate the understanding of our results, we outline the main contributions below with pointers:
 \squishlist
    \item We extend an analytical framework to quantify the effects of memorizing noisy labels (Theorem \ref{thm:main:tau}~-~\ref{thm:noisy}).
    \item We highlight the scenarios when existing popular robust learning methods succeed or fail at the instance level (Section~\ref{subsec:paradox}~\&~\ref{sec:analysis}). We provide the conditions under which the existing approaches improve over memorizing the noisy labels (Theorem~\ref{thm:losscorrection}~\&~\ref{thm:peerloss} and their corollaries), and when not (Theorem~\ref{thm:losscorrection:lower}~\&~\ref{thm:peerloss:lower} and their corollaries).
    \item Our results in Section~\ref{sec:analysis} help explain some empirical observations reported in the literature, including i) peer loss \cite{liu2019peer} induces confident prediction (Lemma~\ref{lemma:peerloss}), and ii) when peer loss and label smoothing \cite{lukasik2020does} could perform better than loss correction \cite{natarajan2013learning,Patrini_2017_CVPR}, which uses explicit knowledge of the noise rates (Section~\ref{sec:ls}~\&~\ref{sec:pl}) - in contrast, peer loss does not require this knowledge. 
    \squishend
Due to space limit, all proofs can be found in the Appendix.
\subsection{Related works}

There have been substantial discussions on the memorization effects of deep neural networks, and how memorization relates to generalization \cite{zhang2016understanding,neyshabur2017exploring,arpit2017closer,feldman2020does,feldman2020neural}. Most relevant to us, recent works have reported negative consequences of memorizing noisy labels, and have proposed corresponding fixes \cite{natarajan2013learning,liu2016classification,liu2019peer,song2019prestopping,yao2020searching,cheng2017learning,pmlr-v97-chen19g,han2020sigua,song2019prestopping}. Different solutions seem to be effective when guarding different type of noise, but there lacks a unified framework to understand why one approach would work and when they would fail. 

Recently, there is increasing attention on learning with instance-dependent noise, which proves to be a much more challenging case \cite{cheng2017learning,cheng2020learning,xia2020parts,zhu2020second}. Our work echoes this effort and emphasizes the instance-level understanding. This focus particularly suits a study with long-tail distributions of instances that appear with different frequencies, which is often shown to be the case with image datasets \cite{zhu2014capturing}.  


\textbf{Common solutions} toward learning with noisy labels build around loss or label corrections \cite{natarajan2013learning,Patrini_2017_CVPR,xia2019anchor}. More recently, light and easy-to-implement solutions are proposed too \cite{lukasik2020does,liu2019peer}. We delve into three of them in Section \ref{sec:fix}. As an area with growing interests, there exist many other solutions - we will not have space to list all, but we want to mention the following two streams of efforts. \emph{Sample cleaning}: Sample cleaning leverages the idea of detecting instance $x$ whose label is corrupted ($\tilde{y} \neq y$) \cite{jiang2017mentornet,han2018co,yu2019does,yao2020searching,wei2020combating,cheng2020learning}. Then the training is mainly done with the selected clean instances, with the aid of processed information from the detected corrupted examples. \emph{Robust loss function}: The literature has also observed the proposal of robust loss functions that perform well with dealing outlier noisy examples \cite{zhang2018generalized,menon2019can,charoenphakdee2019symmetric,wang2019symmetric}. 

\subsection{Overview of the main results: Disparate impacts and treatments of label noise}
Our first set of results, perhaps non-surprisingly, show the disparate impact of noisy labels at the instance level. The impact to the drop of generalization power is linearly dependent on the frequency of the instance and its labels' associated noise rate:
\begin{theorem}[\textbf{Disparate Impacts, Informal}]
For an instance $x$ that appears $l$ times ($l$-appearance instance) in the training data (with $n$ samples), a model $h$ memorizing its $l$ noisy labels leads to the following order of individual excessive generalization error: 
\vspace{-0.05in}
\[
\Omega\left( \frac{l^2}{n^2} \cdot \text{(label noise rate at $x$)}\right)
\]
\vspace{0.05in}
\label{thm:noisy}
\end{theorem}
\vspace{-0.2in}
Our discussions then move to how the existing treatments fare at the instance level. We will introduce a memorization paradox in Section \ref{subsec:paradox} to highlight a common pitfall when analyzing the performance of existing algorithmic treatments at the population level, when a deep model is considered and is able to memorize training examples. Probably more alarmingly, we then provide a set of instance-level analysis to show the disparate treatments of several existing learning with noisy label solutions:
\begin{theorem}[\textbf{Disparate Treatments, Informal}]
For an instance $x$ that appears $l$ times in the training data ($l$-appearance instance), \textbf{when $l$ is large (high-frequency instance)}, with high probability, performing loss correction \cite{natarajan2013learning,Patrini_2017_CVPR} and using peer loss correction \cite{liu2019peer} on $x$ improves generalization performance compared to memorizing the noisy labels. \textbf{When $l$ is small (long-tail instance)}, with non-negligible probability, both loss correction and peer loss incur higher prediction errors than memorizing the noisy labels. 
\end{theorem}

\section{Formulation}\label{sec:formulation}
To reuse the main analytical framework built in \citet{feldman2020does}, we largely follow their notations. In the clean setting, a training dataset $S=\{(x_1,y_1),...,(x_n,y_n)\}$ is available. Each $x$ indicates a feature vector and each $y$ is an associated label.  Denote by $X$ the space of $x$ and $Y$ the space for $y$. Jointly $(x,y)$ are drawn from an unknown distribution $\Pd$ over $X \times Y$. Specifically, $x$ is sampled from a distribution $\D$, and the true label $y$ for $x$ is specified by a function $f: X \rightarrow Y$ \emph{drawn from a distribution $\F$}.

The learner's algorithm $\A$, as a function of the training data $S$, returns a distribution of classifiers or functions $h: X  \rightarrow Y$. By this, we consider a randomized algorithm that would potentially lead to the deployment of a randomized classifier. We define the following generalization error
$
\err_{\Pd}(\A,S):= \E_{h \sim \A(S)}[\err_\Pd(h)],
$
where $\err_{\Pd}(h):= \E_{\Pd}[\mathds{1}(h(x) \neq y)]$ and $\mathds{1}(\cdot)$ is the indicator function.
When there is no confusion, we shall use $x,y$ to denote the random variables generating these quantities when used in a probability measure. To better and clearly demonstrate the main message of this paper, we consider discrete domains of $X$ and $Y$ such that $|X|=N,|Y|=m$. Our model, as well as the main generalization results, can mostly extend to a setting with continuous $X$ (Section 4, \cite{feldman2020does}). We briefly discuss it after we introduce the following process to capture the generation of each instance $x$: We follow  \citet{feldman2020does} to characterize an unstructured discrete domain of classification problems:
\squishlist
    \item Let $\pi = \{\pi_1,...,\pi_N\}$ denote the priors for each $x \in X$.
    \item For each $x \in X$, sample a quantity $p_x$ independently and uniformly from the set $\pi$.
    \item Then the resulting probability mass function of $x$ is given by $D(x)=\frac{p_x}{\sum_{x \in X} p_x}$ - this forms the distribution $\mathcal D$ that $x$ will be drawn from. 
\squishend
For the case with continuous $X$, instead of assuming a prior $\pi$ over each $x$ in a finite $X$, it is assumed there is a prior $\pi$ defined over $N$ mixture models. Each $x$ has a certain probability of being drawn from each model and then will realize according to the generative model. Each of the generative models captures similar but non-identical examples. With the above generation process, denote by $\p[\cdot |S]$ the marginal distribution over $\Pd$ conditional on $S$, we further define the following conditional generalization error (on the realization of the training data $S$):
\[
\err(\pi,\F, \A|S) := \E_{\Pd \sim \p[\cdot |S]}\left[\err_{\Pd}(\A,S)\right].
\]
\textbf{$l$-appearance instances}:
We denote by $X_{S=l}$ the set of $x$s that appeared exactly $l \geq 1$ times in the dataset $S$. The difference in $l$ helps us capture the imbalance of the distribution of instances. Later we show that the handling of instances with different frequencies matters differently. 

\subsection{Noisy labels}
We consider a setting where the training labels are noisy. Suppose for each training instance $(x,y)$, instead of observing the true label $y$, we observe a noisy copy of it, denoting by $\tilde{y}$. Each $\tilde{y}$ is generated according to the following model:
\begin{align}
    T_{k,k'}(x) := \p[\tilde{y} = k' |y=k, x], k',k \in Y.
\end{align}
We will denote by $T(x) \in \R^{m \times m}$ the noise transition matrix with the $(k,k')$-entry defined by $ T_{k,k'}(x)$.
Each of the above noisy label generation is independent across different $x$. We have access to the above noisy dataset $S':=\{(x_1,\tilde{y}_1),...,(x_n,\tilde{y}_n)\}$.

\begin{figure}[!t]
\centering
\includegraphics[width=0.45\textwidth]{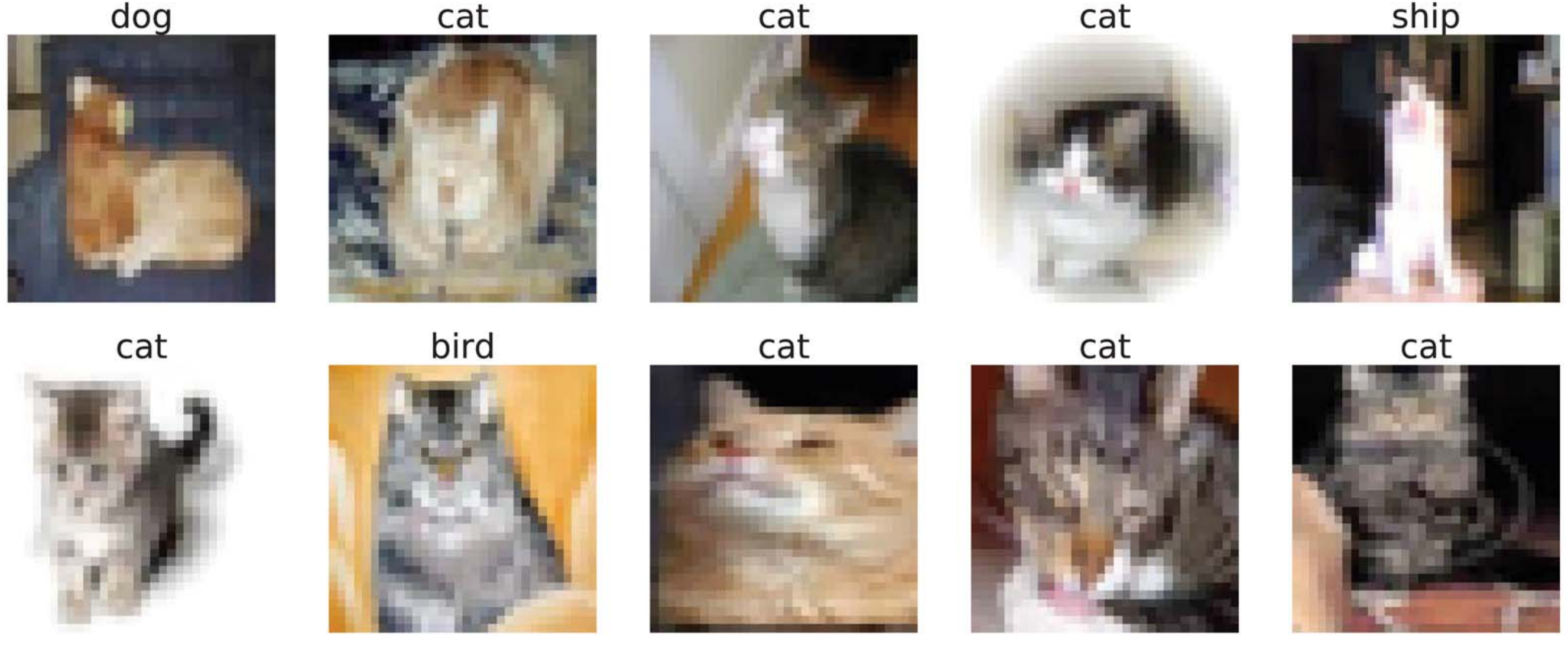}
\caption{Sample examples of ``Cats" in CIFAR-10  with noisy labels on top. Examples are taken from \cite{zhu2021clusterability}.}
\label{fig:cifar}
\end{figure}

 An $x$ that appears $l$ times in the dataset will have $l$ independently generated noisy labels. One can think of these as $l$ similar data instances, with each of them equipped with a single noisy label collected independently. 
 For instance, Figure \ref{fig:cifar} shows a collection of 10 similar ``Cats" from the CIFAR-10 dataset \cite{krizhevsky2009learning}. On top of each image, we show a ``noisy" label collected from Amazon Mechanical Turk. Approximately, one can view each row as a $x$ with 5 appearances, with each of the instances associated with a potentially corrupted label.  
    
 $T(x)$ varies across the dataset $S$, and possibly that $T(x)$ would even be higher for low-frequency/rare instances, due to the inherent difficulties in recognizing and labeling them. 

Note the noisy label distribution $\p[\tilde{y}|x]$ has a larger entropy due to the additional randomness introduced by $T(x)$ and is therefore harder to fit. As we shall see later, this fact poses additional challenges, especially for the long-tail instances that have an insufficient number of observations. 

\section{Impacts of Memorizing Noisy Labels}\label{sec:noise}
In this section, we discuss the impacts of noisy labels when training a model that can memorize examples. Our analysis builds on a recent generalization bound for studying the memorization effects of an over-parameterized model. 

\subsection{Generalization error}
Denote by $\bar{\pi}^N$ the resulting marginal distribution over $x$: $\bar{\pi}^N(\alpha):=\p[D(x)=\alpha]$. $\bar{\pi}^N$ controls the true frequency of generating instances. Define the following quantity:
\begin{align}
    \tau_l := \frac{\E_{\alpha \sim \bar{\pi}^N}[\alpha^{l+1}\cdot (1-\alpha)^{n-l}]}{\E_{\alpha \sim \bar{\pi}^N}[\alpha^{l}\cdot (1-\alpha)^{n-l}]}.
\end{align}
Intuitively, $\tau_l$ quantifies the ``importance weight" of the $l$-appearance instances. Theorem~2.3 of \citet{feldman2020does} provides the following generalization error of an algorithm $\A$: 
\begin{theorem} [Theorem 2.3, \cite{feldman2020does}]
For every learning algorithm $\A$ and every dataset $S \in (X \times Y)^n$:
\begin{align}
    &\err(\pi,\F, \A|S) \geq opt(\pi,\F|S)\nonumber \\
    &+\sum_{l \in [n]} \tau_l \cdot \sum_{x \in X_{S=l} }\p_{h \sim \A}[h(x) \neq y],
\end{align}
where in above, $opt(\pi,\F|S) := \min_{\A} \err(\pi,\F,\A|S)$ is the minimum achievable generalization error.
\end{theorem}
We will build our results and discussions using this generalization bound. Our discussion will focus on how label noise can disrupt the training of a model through the changes of the following \textbf{Excessive Generalization Error:}
\begin{align}
\err^{+}(\Pd,\A|S):=&\sum_{l \in [n]} \tau_l \sum_{x \in X_{S=l} }\p_{h \sim \A(S')}[h(x) \neq y], \label{eqn:excesserror}
\end{align}
Note though the input of the algorithm $\A$ is the noisy dataset $S'$, we are interested in the distribution conditional on the clean dataset $S$ - this is the true distribution that we aim for $h$ to generalize to. On the other hand, the distribution induced by $S'$ will necessarily encode bias to the clean distribution that we are interested in, when some labels are indeed different from the true ones. Even though we do not have access to $S$, the above ``true generalization error'' is well-defined for our analysis, and nicely encodes three quantities that are of primary interests to our study:
\squishlist
    \item $\tau_l$: the ``importance weight" of the $l$-appearance instances. 
        \item $l$: the frequency of instances  that categorizes how popular a particular instance $x$ is in the dataset.
    \item $\sum_{x \in X_{Z=l} }\p_{h \sim \A(S')}[h(x) \neq y]$: the accumulative generalization error $h$ makes for $l$-appearance instances.
\squishend

We will also denote by 
\begin{align}
\err^{+}_l(\Pd,\A, x|S):=& \tau_l \cdot \p_{h \sim \A(S')}[h(x) \neq y] \label{eqn:excesserror:ind}
\end{align}
the \textbf{Individual Excessive Generalization Error} caused by a $x \in X_{S=l}$. Easy to see that $\err^{+}(\Pd,\A|S) = \sum_x \err^{+}_l(\Pd,\A, x|S)$.

\subsection{Importance of memorizing  an $l$-appearance instance}
Clearly Eqn.~(\ref{eqn:excesserror}) informs us that different instance contributes differently to the generalization error. It was proved in \citet{feldman2020does} even a single-appearance instance $x \in X_{S=1}$ (i.e., $l=1$) will contribute to the increase of generalization error at the order of $\Omega(\frac{1}{n})$: when $\pi_{\max }:= \max_{j \in [N]} \pi_j  \leq 1/200$, we have 
\[
\tau_1 \geq \frac{1}{7n}\cdot \textsf{weight}\left(\pi,\left[\frac{1}{2n},\frac{1}{n}\right]\right),
\]
 where $\textsf{weight}\bigl(\pi,\bigl[\beta_1,\beta_2\bigr]\bigr)$ is the expected fraction of distribution $D$ contributed by frequencies in the range $[\beta_1,\beta_2]$:
 \begin{align*}
 \textsf{weight}\bigl(\pi,\bigl[\beta_1,\beta_2\bigr]\bigr):= \E\left[\sum_{x \in X}D(x) \cdot \mathds{1}\left(D(x) \in [\beta_1,\beta_2]\right)\right]
\end{align*}
The expectation is w.r.t. $D(x) \sim \pi$ (and followed by the normalization procedure).
 We next first generalize the above lower bound to $\tau_l$ for an arbitrary $l$:
 \begin{theorem}
 For sufficiently large $n,N$, when $\pi_{max} \leq \frac{1}{20}:$
 \begin{align}
     \tau_l \geq 0.4 \cdot \frac{l (l-1)}{n(n-1)} \cdot \textsf{weight}\left(\pi,\left[\frac{2}{3}\frac{l-1}{n-1}, \frac{4}{3}\frac{l}{n}\right]\right)
 \end{align}
 \label{thm:main:tau}
 \end{theorem}
 
We observe that $\frac{l (l-1)}{n(n-1)} = O(\frac{l^2}{n^2})$. For instance: 
 \squishlist
     \item  An $l = O(n^{2/3})$-appearance instance will lead to an $\Omega(\frac{1}{n^{2/3}})$ order of impact.
     \item An $l = O(n^{3/4})$-appearance instance will lead to an $\Omega(\frac{1}{\sqrt{n}})$ order of impact.
     \item An $l = c n$-appearance (linear) instance will lead to an $\Omega(1)$ bound, a constant order of impact.
 \squishend
 Secondly, for the $\textsf{weight}\bigl(\pi,\bigl[\frac{2}{3}\frac{l-1}{n-1}, \frac{4}{3}\frac{l}{n}\bigr]\bigr)$ term, we have the frequency interval at the order of length
$
     \frac{l}{n} - \frac{l-1}{n-1} = \frac{n-l}{n(n-1)} = O(\frac{1}{n}).
$
That is the weight term captures the frequency of an $O(\frac{1}{n})$ interval of the sample distribution. One might notice that there seems to be a disagreement with the reported result in \citet{feldman2020does} when $l$ is small (particularly when $l=1$): when ignoring the $\textsf{weight}$ term $\tau_l$, an $O(\frac{1}{n})$ lower bound was reported, while ours leads to an $ O(\frac{1}{n^2})$ one. This is primarily due to different bounding techniques we incurred.  Our above bound suits the study of $l$ that is on a higher order than $O(\frac{1}{n})$. For small $l$, we provide the following bound:
\begin{theorem}
 For sufficiently large $n,N$ and $\pi_{max} \leq \frac{1}{20}:$
 \begin{align*}
     \tau_l \geq 0.4  \frac{l-1}{n-1} \cdot \frac{1}{1.1^l} \cdot \textsf{weight}\left(\pi,\left[0.7  \frac{l-1}{n-1}, \frac{4}{3}\frac{l-1}{n-1}\right]\right)
 \end{align*} 
 \label{prop:tau}
\end{theorem}
\vspace{-0.2in}
  When $l$ is small, based on the above bound, we do see $\tau_l = \Omega(\frac{1}{n})$, while the $\textsf{weight}$ constant again captures an $O(\frac{1}{n})$ interval of instances. Note that this bound becomes less informative as $l$ grows, due to the increasing $1.1^l$ term. Also when $l=1$, our bound becomes vacuous since $l-1=0$. 
 
 \subsection{Memorizing noisy labels}
 
\begin{figure}[!ht]
\centering
\includegraphics[width=0.23\textwidth]{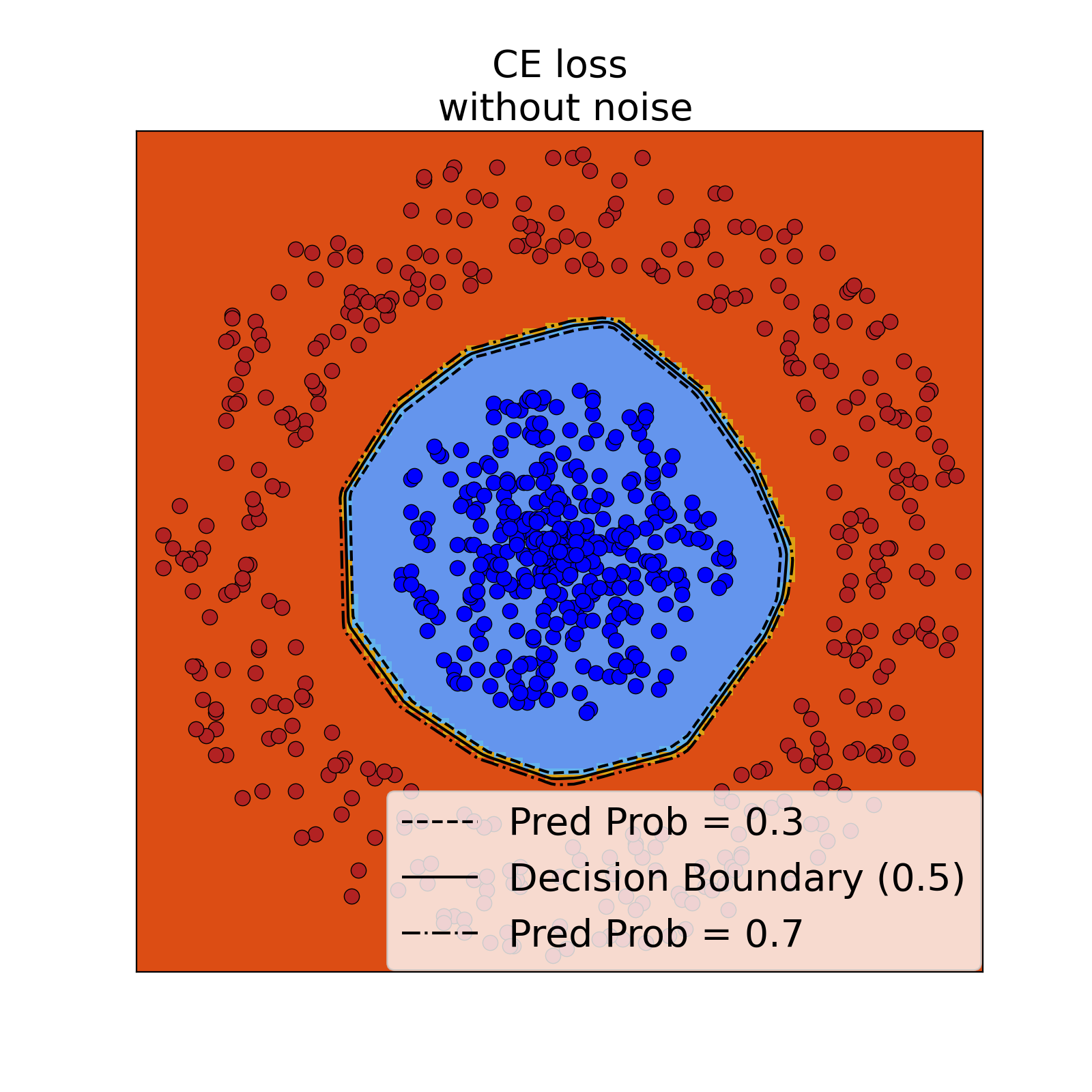}\hspace{-0.2in}
\includegraphics[width=0.23\textwidth]{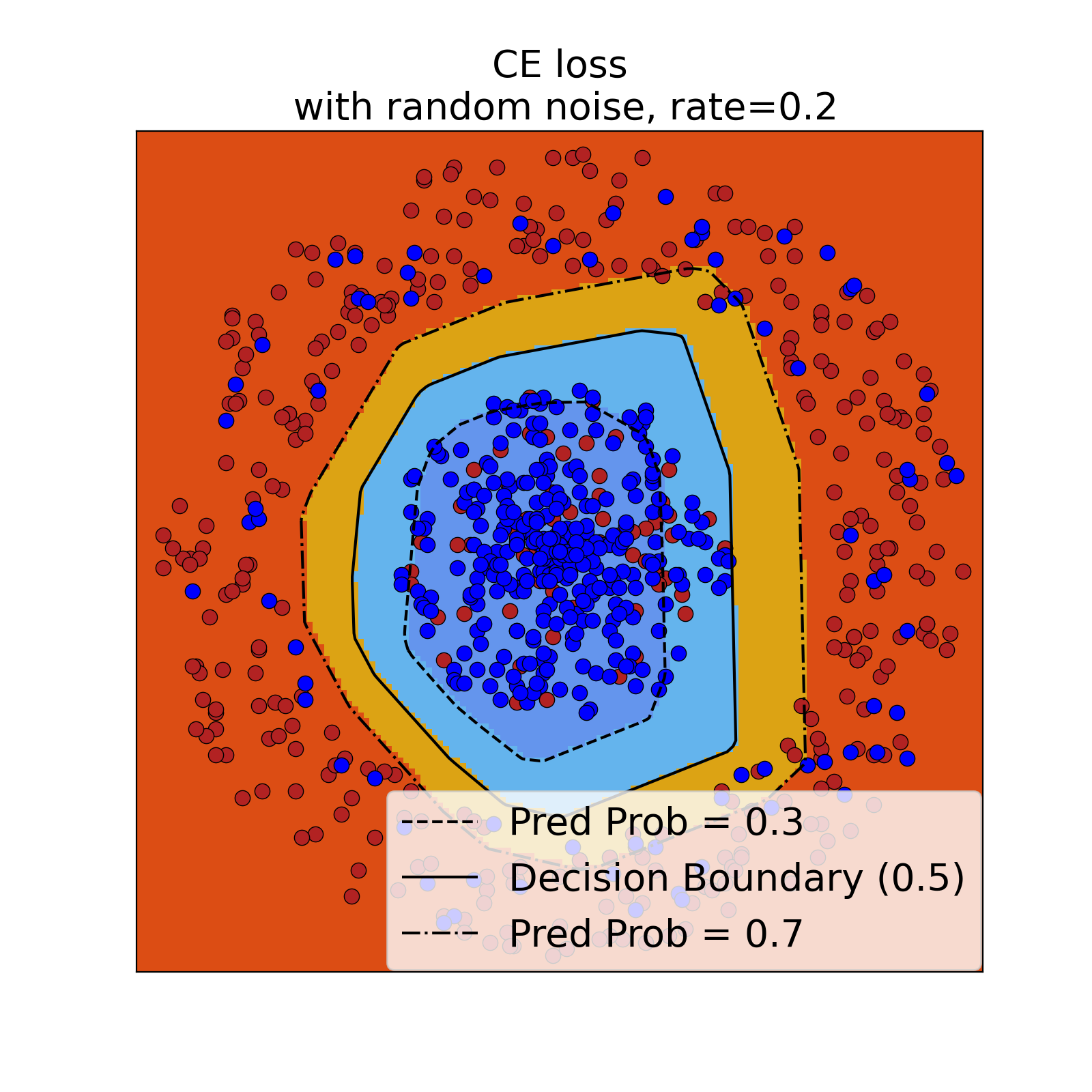}
\vspace{-0.2in}

\caption{A 2D example illustrating the memorization of noisy labels. Left panel: Training with clean labels. Right panel: Training with random $20\%$ noisy labels. Example with $40\%$ noisy labels can be found in the Appendix.} 
\label{fig:2d:ce}
\end{figure}

In order to study the negative effects of memorizing noisy labels, we first define the memorization of noisy labels. For an $x \in X_{S=l}$ and its associated $l$ noisy labels, denote by $\tilde{\p}[\tilde{y}=k|x], k \in Y$ the empirical distribution of the $l$ noisy labels: for instance when $l=3$ and two noisy labels are 1, we have $\tilde{\p}[\tilde{y}=1|x]=\frac{2}{3}$.
\begin{definition}[Memorization of noisy labels] We call a model $h$ memorizing noisy labels for instance $x$ if $\p_{h \sim \A(S')}[h(x) = k] = \tilde{\p}[\tilde{y}=k|x]$. \label{def:memnoisy}
\end{definition}
Note that the probability measure is over the randomness of the algorithm $\A$, as well as the potential randomness in $h$ - practically one can sample a classification outcome based on the posterior prediction of $h(x)$.
Effectively the assumption states that when a model, e.g. a deep neural network, memorizes all $l$ noisy labels for instance $x$, its output will follow the same empirical distribution. It has been shown in the literature \cite{cheng2017learning,cheng2020learning} that a fully memorizing neural network will be able to  encode $\tilde{\p}[\tilde{y}=k|x]$ for each $x$. This is also what we observe empirically. In Figure \ref{fig:2d:ce}, we simulate a 2D example: there are two classes of instances. The outer annulus represents one class and the inner ball is the other. Given the plotted training data, we train a 2-layer neural network using the cross-entropy (CE) loss. On the left panel, we observe concentrated predictions from the trained model when labels are clean. However, the decision boundary (colored bands with different prediction probabilities) becomes less certain and more probabilistic with the addition of noisy labels, signaling that the neural network is memorizing a mixed distribution of noisy labels. 
\vspace{-0.2in}

\vspace{-0.1in}

\begin{figure}[!ht]
\centering
\includegraphics[width=0.48\textwidth]{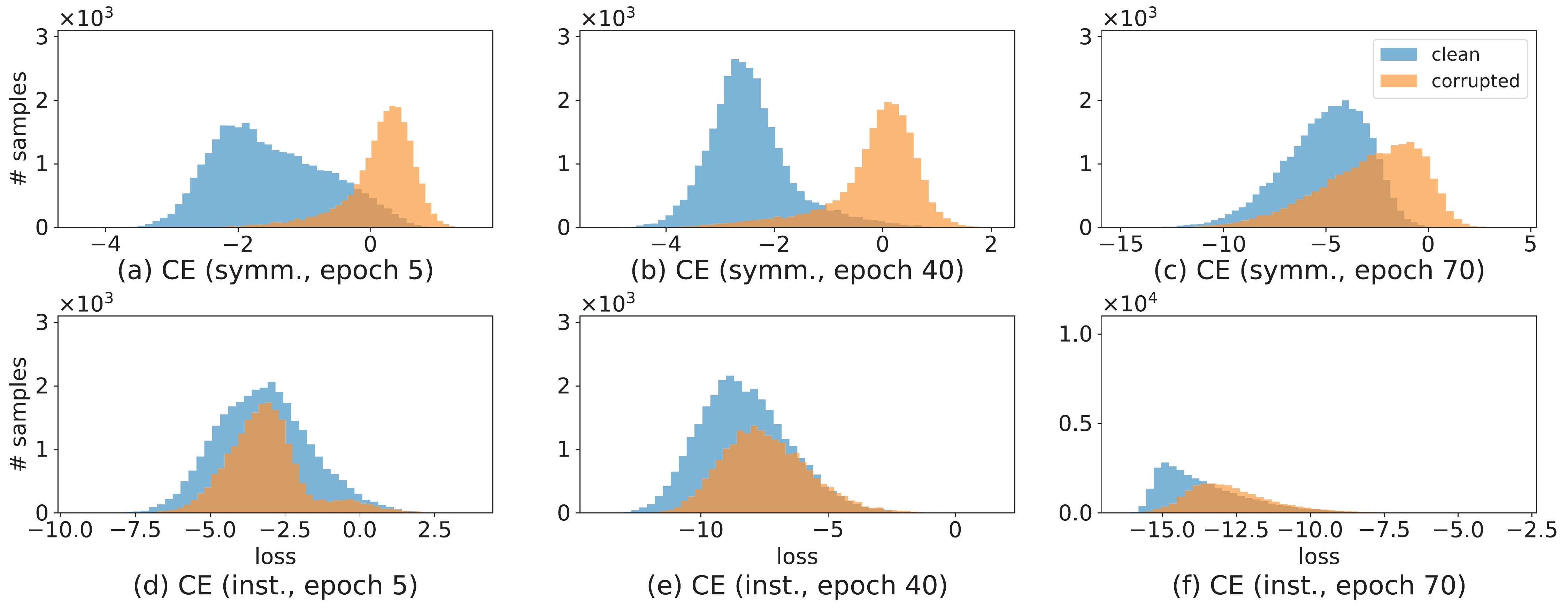}  
\vspace{-0.2in}
\caption{Memorization effects on CIFAR-10 with noisy labels.}
\label{fig:cifar-memorization}
\end{figure}

We further illustrate this in Figure \ref{fig:cifar-memorization} where we train a neural network on the CIFAR-10 dataset with synthesized noisy labels. The top row simulated simple cases with instance-independent noise $T(x) \equiv T$, while the bottom one synthesized an instance-dependent case\footnote{We defer the empirical details to the Appendix.}. In each row, from Left to Right, we show the progressive changes in the distribution of losses\footnote{To better visualize the separation of the instances, we follow \citet{cheng2020learning} to plot the distribution of a normalized loss by subtracting the CE loss with a normalization term $\sum_{k} \p[h(x)=k]/m$, resulting possibly negative losses on $x$-axis. } across different training epochs. Preferably, we would like the training to return two distributions of losses that are less overlapped so the model can better distinguish the clean (colored in {\color{blue}blue}) and corrupted instances (colored in {\color{orange}orange}). 
However, we do observe that in both cases, the neural network fails to separate the clean instances from the corrupted ones and memorizes a mixture of both.

This definition of memorization is certainly a simplification but it succinctly characterizes the situation when there are $l$ similar but non-identical instances, the deep neural network would memorize the noisy label for each of them, which then results in memorizing each realized noisy label class a $\tilde{\p}[\tilde{y}=k|x]$ fraction of times. This definition would also require the instances ($x$'s) to be rather independent, or each $x$'s own label information is the most dominant one, which is likely to be true when $N$ is large enough to separate $X$. Most of our observations would remain true as long as the memorization leads $h$ to predict in the same direction of $\tilde{\p}[\tilde{y}=k|x]$. In particular, when $h$ does not fully remember the empirical label distribution, we conjecture that our main results hold if the memorization preserves orders: for any two classes $k,k'$, if $\tilde{\mathbb P}[\tilde{y}=k|x]>\tilde{\mathbb P}[\tilde{y}=k'|x]$, we require $h$ to satisfy $\mathbb P_{h  \sim \mathcal A(S')}[h(x)=k]>\mathbb P_{h  \sim \mathcal A(S')}[h(x)=k']$. This simplification in Definition \ref{def:memnoisy} greatly enables a clear presentation of our later analysis.

 \subsection{Impacts of memorizing noisy labels}
 
Based on Theorem \ref{thm:main:tau}, we summarize our first observation that over-memorizing noisy labels for higher frequency instances leads to a bigger drop in the generalization power:
\begin{theorem}
For $x \in X_{S=l}$ 
with true label $y$, $h$ memorizing its $l$ noisy labels leads to the following order of individual excessive generalization error $\err^{+}_l(\Pd,\A, x|S)$: 
\vspace{-0.1in}
\[
\Omega\left( \frac{l^2}{n^2} \cdot \textsf{weight}\left(\pi,\left[\frac{2}{3}\frac{l-1}{n-1}, \frac{4}{3}\frac{l}{n}\right] \right)\cdot \sum_{k \neq y} \tilde{\p}[\tilde{y}=k|x]\right)
\]
\label{thm:noisy}
\end{theorem}
\vspace{-0.2in}
We would like to note that with large $l$, $\sum_{k \neq y}\tilde{\p}[\tilde{y}=k|x] \rightarrow \sum_{k \neq y}T_{y,k}(x)$ - not surprisingly, the higher probability an instance is observing a corrupted label, the higher generalization error it will incur.
The bound informs us that over-memorizing high-frequency instances lead to a larger negative impact on the generalization. However, we shall see later the higher-frequency instances are in fact the easier ones to fix!
On the other hand, memorizing the noisy labels for the lower frequency/appearance instances leads to a smaller drop in generalization performance. Nonetheless, they do incur non-negligible changes. For instance, misremembering a single instance with $l=1$ leads to an $O(\frac{1}{n})$ increase in generalization error. Later we show a small $l$ poses additional challenges in correcting the mistakes.  

\section{Learning with Noisy Labels}\label{sec:fix}

In this section, we quickly review a subset of popular and recently proposed solutions for learning with noisy labels. 

\subsection{Loss correction}

Arguably one of the most popular approaches for correcting the effects of label noise is through loss correction using the knowledge of $T(x)$ \cite{natarajan2013learning,liu2016classification,Patrini_2017_CVPR}. Let's denote by $\ell: \R^{m} \times Y \rightarrow \R_{+}$ the underlying loss function we adopted for training a deep neural network. Denote by $\ell(h(x),y)$ the loss incurred by $h$ on instance $(x,y)$, and $\Bell(h(x))=[\ell(h(x),y')]_{y' \in Y}$ the column vector form of the loss. We will assume each $T(x)$ is invertible that $T^{-1}(x)$ well exists. Loss correction is done via defining a surrogate loss function $\tilde{\ell}$ as follows:
\begin{align}
    \Bell_{\text{LC}}(h(x))= T^{-1}(x) \cdot \Bell(h(x)).\label{eqn:losscorrection}
\end{align}
The reason for performing the above correction is due to the following established unbiasedness property: Denote by $\ny$ the one-hot encoding column vector form of the noisy label $\tilde{y}$: $\ny:=[0;...;\underbrace{1}_{\tilde{y}'s~\text{position}};...;0]$, we have:
\begin{lemma}[Unbiasedness of $ \ell_{\text{LC}}$, \cite{natarajan2013learning}]
$\E_{\tilde{y}|y}[\ny^{\top} \cdot \Bell_{\text{\emph LC}}(h(x))] = \ell(h(x),y)$.
\label{lemma:unbias}
\end{lemma}
The above lemma states that when conditioning on the distribution of $\tilde{y}|y$, $\Bell_{\text{LC}}(h(x)) $ is unbiased in expectation w.r.t. the true loss $\ell(h(x),y)$. In Section \ref{subsec:paradox}, we explain how this unbiasedness is established for the binary classification setting.  Based on Lemma \ref{lemma:unbias}, one can perform empirical risk minimization over $\sum_{i=1}^n \ny^{\top}_i \cdot\Bell_{\text{LC}}(h(x_i))$, hoping the empirical sum will approximately converge to its expectation which will then equalize to the true empirical loss $\sum_{i=1}^n \ell(h(x_i),y_i)$. Of course, a commonly made assumption/requirement when applying this approach is that $T(x) \equiv T, \forall x$, and $T$ can be estimated accurately enough. There exist empirical and extensive discussions on how to do so \cite{Patrini_2017_CVPR,xia2019anchor,yao2020dual,zhang2021learning,li2021provably}.

\subsection{Label smoothing}

Label smoothing has demonstrated its benefits in improving learning representation \cite{muller2019does}. A recent paper \cite{lukasik2020does} has also proved the potential of label smoothing in defending training against label noise. Denote by $\1$ the all-one vector, and a smoothed and soft label is defined as 
$
    \sy := (1-a) \cdot \ny + \frac{a}{m} \cdot \1,
$
where $a \in [0,1]$ is a smoothing parameter. 
That is, $\sy$ is defined as a linear combination of the noisy label $\tilde{y}$ and an uninformative and uniform label vector $\1$. Then each instance $x$ will be evaluated using $\sy$: 
$
\sy^{\top} \cdot \Bell(h(x))
.
$

Though label smoothing has shown promising advantages over loss correction, there are few theoretical understandings of why so, except for its high-level idea of being ``conservative" when handling noisy labels.

\subsection{Peer loss}
Peer loss \cite{liu2019peer} is a different line of solution that promotes the use of multiple instances simultaneously while evaluating a particular noisy instance $(x,\tilde{y})$. A salient feature of peer loss is that the implementation of it does not require the knowledge of $T(x)$. The definition for peer loss has the following key steps: 
\squishlist
    \item For each $(x,\tilde{y})$ we aim to evaluate, randomly drawn two other sample indices $p_1, p_2 \in [n]$.
    \item  Pair $x_{p_1}$ with $\tilde{y}_{p_2}$, define peer loss:
    \vspace{-0.05in}
\begin{align}
    \ell_{\text{PL}}(h(x),\tilde{y}) := \ell(h(x),\tilde{y}) - \ell(h(x_{p_1}),\tilde{y}_{p_2}).\label{eqn:peerloss}
\end{align}
\squishend
\vspace{-0.05in}
When $T(x) \equiv T$, it was proved in \cite{liu2019peer} that for binary classification with equal label prior, when $\ell$ is the 0-1 loss, minimizing peer loss returns the same minimizer of $\E[\ell(h(x),y)]$ on the clean distribution.

\subsection{Memorization paradox}\label{subsec:paradox}

Some of the above approaches have established strong theoretical guarantees of recovering the optimal classifier in expectation when using only noisy training labels \cite{natarajan2013learning,liu2019peer,ma2020normalized}. Why would we need a different understanding? First of all, most theoretical results assumed away the outstanding challenges of having an unknown number and distribution of noise rate matrix $T(x)$ (either needed for estimation purpose or for handling them implicitly) and focused on a single transition matrix $T$. Secondly, the existing error analysis often focuses on the distribution level, while we would like to zoom in to each instance that occurs with a different frequency.

In addition, we now highlight a paradox introduced by a commonly made assumption that the noisy labels and the model's prediction $t$ are conditionally independent given true label $y$:
$
\p[t,\tilde{y}|y] = \p[t|y]\cdot \p[\tilde{y}|y]
$, or that $t$ is simply deterministic that it does not encode the information of $\tilde{y}$. This assumption is often needed when evaluating the expected generalization error under noisy distributions. The independence can be justified by modeling $\tilde{y}$ as being conditionally independent of feature $x$, which the prediction $t$ is primarily based on. 

Let's take loss correction for an example. For a clear demonstration, let's focus on the binary case $y \in \{-1,+1\}$. Consider a particular $x$, define 
$
    e_-(x) := \p[\tilde{y}=+1|y=-1,x],~e_+(x) := \p[\tilde{y}=-1|y=+1,x]
$ and $T(x)$:
\begin{align}
T(x):=\begin{bmatrix}
1-e_-(x) & e_-(x) \\
e_+(x) & 1-e_+(x) 
\end{bmatrix}
\label{eqn:T}    
\end{align}
Easy to verify its inverse is: 
\begin{align}
T^{-1}(x) = \frac{1}{1-e_+(x)-e_-(x)} 
\begin{bmatrix}
1-e_+(x) & -e_-(x)\\
 -e_+(x) & 1-e_-(x)
\end{bmatrix}
\label{eqn:T:inverse}
\end{align}
For the rest of this section, without confusion, let's shorthand $e_+(x),e_-(x)$ as $e_+,e_-$.
Then loss correction (Eqn. (\ref{eqn:losscorrection})) takes the following form: 
\begin{small}
\begin{align*}
    \ell_{\text{LC}}(h(x),-1)=\frac{(1-e_+)\cdot \ell(h(x),-1)-e_- \cdot \ell(h(x),+1)}{1-e_+-e_-}\\
    \ell_{\text{LC}}(h(x),+1)=\frac{(1-e_-)\cdot \ell(h(x),+1)-e_+ \cdot \ell(h(x),-1)}{1-e_+-e_-}
\end{align*}
\end{small}
Consider the case with true label $y=+1$. The following argument establishes the unbiasedness of $\ell_{\text{LC}}$ (reproduced from \citet{natarajan2013learning}, with replacing and instantiating a prediction $t$ with $h(x)$):
\vspace{-2pt}
\begin{tcolorbox}[colback=grey!5!white,colframe=grey!5!white]
\vspace{-10pt}
\begin{small}
\begin{align*}
\hspace{-0.2in}
    &\E_{\tilde{y}|y=+1}[\ell_{\text{LC}}(h(x),\tilde{y})]\\
    &= (1-e_+) \cdot \ell_{\text{LC}}(h(x),+1) + e_+ \cdot \ell_{\text{LC}}(h(x),-1) \\
    &=(1-e_+)  \frac{(1-e_+)\cdot \ell(h(x),-1)-e_- \cdot \ell(h(x),+1)}{1-e_+-e_-}\\
    &+ e_+  \frac{(1-e_-)\cdot \ell(h(x),+1)-e_+\cdot \ell(h(x),-1)}{1-e_+-e_-}\\
    &=\ell(h(x),+1) = \ell(h(x),y=+1)~,
\end{align*}
\end{small}
\vspace{-18pt}
\end{tcolorbox}
\vspace{-2pt}
That is the conditional expectation of $\ell_{\text{LC}}(h(x),\tilde{y})$ recovers the true loss $\ell(h(x),y)$. Nonetheless, the first equality assumed the conditional independence between $h$ and $\tilde{y}$, given $y$. When $h$ is output from a deep neural network and memorizes all noisy labels $\tilde{y}$, the above independence condition can be challenged. As a consequence, it is unclear whether the classifiers that fully memorize the noisy labels would result in a lower empirical loss during training. We call the above observation the \emph{memorization paradox}. We conjecture that this paradox leads to inconsistencies in previously observed empirical evidence, especially when training a deep neural network solution that memorizes labels well. In the next section, we will offer new explanations for how the proposed solutions actually fared when the trained neural network is able to memorize the examples.

\section{How Do Solutions Fare at Instance Level?}\label{sec:analysis}
In this section, we revisit how the above solutions offer fixes at the instance level and under what conditions they might fail to work. Unless stated otherwise, throughout the section, we focus on a particular instance $x \in X_{S=l}$ with true label $y$ and $l$ corresponding noisy labels $\tilde{y}$'s, one for each appearance. With limited space, the goal is to provide a template for carrying out further analysis for methods that are of individual interest. 
The three presented approaches were selected carefully as representatives for:
\squishlist
    \item Mainstream approaches (loss correction, label correction, loss reweighting etc) that use noise transition matrix $T$ (\textbf{loss correction} in this paper).
    \item Robust losses that regularize against noisy outliers (\textbf{label smoothing} in this paper).
    \item More recent approaches that do not require the noise transition matrix (\textbf{peer loss} in this paper).
\squishend

\subsection{Loss correction}

We start with loss correction and notice that the loss correction step is equivalent to the following ``label correction" \footnote{Please note that our `` label correction" definition differs from the existing ones in the literature.} procedure. Denote by $\p(\ny):=[\p[\tilde{y}=k|x]]_{k \in Y}$ the vector form of the distribution of noisy label $\tilde{y}$, and $\y$ as the vector form of one-hot encoding of the true label $y$. As assumed earlier, the generation of noisy label $\tilde{y}$ follows: 
$
\p(\ny) := T^{\top}(x) \cdot \y.
$
When $T(x)$ is invertible (commonly assumed), we will have
$
\y = (T^{-1}(x))^{\top}\cdot \p(\ny)
$, that is when $\p(\ny)$ is the true and exact posterior distribution of $\tilde{y}$, $ (T^{-1}(x))^{\top}\cdot \p(\ny)$ recovers $\y$.
Based on the above observation, easily we can show that (using linearity of expectation) loss correction effectively pushed $h$ to memorize $(T^{-1}(x))^{\top}\cdot \p(\ny)=\y$, i.e. the clean label:
\vspace{-5pt}
\begin{tcolorbox}[colback=grey!5!white,colframe=grey!5!white]
\vspace{-15pt}
\begin{align}
&\E_{\tilde{y}|y}[\ny^{\top} \cdot \Bell_{\text{LC}}(h(x))] \nonumber \\
=&\E_{\tilde{y}|y}[\ny^{\top} \cdot T^{-1}(x) \cdot \Bell(h(x))] \nonumber\\
=&
     \E_{\mathbf{y}' \sim  (T^{-1}(x))^{\top}\cdot \p(\ny)}\left[(\mathbf{y}')^{\top} \cdot \Bell(h(x))\right] \nonumber \\
     =& \y^{\top}\cdot \Bell(h(x))
     \label{eqn:lc}
\end{align}
\vspace{-25pt}
\end{tcolorbox}
\vspace{-2pt}
That is loss correction encourages $h$ to memorize the true label $\y$, therefore reducing $\p[h(x)\neq y]$ to 0 to improve generalization. 

The above is a clean case with accessing $\p(\ny)$, the exact noisy label distribution, which differs from the empirical noisy label distribution $\tilde{\p}[\tilde{y}=k|x]$ that a deep neural network can access and memorize. This is mainly due to the limited number, $l$, of noisy labels for an $x \in X_{S=l}$. Denote by $\tilde{\p}(\ny)$ the vector form of $\tilde{\p}[\tilde{y}=k|x]$.  Let $\smy $ be the  ``corrected label" following from the distribution defined by $(T^{-1}(x))^{\top}\cdot \tilde{\p}(\ny)$: 
\[
\text{Corrected Label: }~~\smy 
= (T^{-1}(x))^{\top}\cdot \tilde{\p}(\ny).
\]
Denote by $x(1),...,x(l)$ the $l$ appearance of $x$, and $\tilde{y}(1),...,\tilde{y}(l)$ the corresponding noisy labels.
Similar to Eqn. (\ref{eqn:lc}) we can show that:
\vspace{-5pt}
\begin{tcolorbox}[colback=grey!5!white,colframe=grey!5!white]
\vspace{-20pt}
\begin{align}
    &\frac{1}{l}\sum_{i=1}^l\ny^{\top}(i) \cdot \Bell_{\text{LC}}(h(x)) \nonumber \\
=&\E_{\tilde{\p}(\ny)}[\ny^{\top} \cdot T^{-1}(x) \cdot \Bell(h(x))] \nonumber\\
     \smy^{\top} \cdot \Bell(h(x))
     \label{eqn:lc2}
\end{align}
\vspace{-25pt}
\end{tcolorbox}
\vspace{-2pt}
That is, the empirical loss for $x$ with loss correction is equivalent with training using $\smy$! Next we will focus on the binary case: $T(x)$ is fully characterized and determined by $e_+(x),e_-(x)$ (Eqn. \ref{eqn:T}). We will follow the assumption made in the literature that $e_+(x) + e_-(x) < 1$ (noisy labels are at least positively correlating with the true label). Easy to prove that the two entries of $\smy$ add up to 1\footnote{For binary labels $\{-1,+1\}$, the first entry of the vectors corresponds to $-1$ ($\mathbf{y}_{\text{LC}}[1]$), the second for $+1$ ($\mathbf{y}_{\text{LC}}[2]$).}:
\begin{lemma}
$\mathbf{y}_{\text{\emph{LC}}}[1] + \mathbf{y}_{\text{\emph{LC}}}[2] =1$.\label{lemma:smy}
\end{lemma}
\vspace{-0.05in}
However, it is possible that $(T^{-1}(x))^{\top}\cdot \tilde{\p}(\ny)$ is not a valid probability measure, in which case we will simply cap $\smy$ at either $[1;0]$ or $[0;1]$.  Denote by $y_{\text{LC}}$ the random variable drawn according to $\smy$.
 We again call $h$ memorizing $\smy$ if $\p[h(x) = k] = \p[y_{\text{LC}}=k|x], \forall k \in Y$. 
Let's simplify our argument by assuming the following equivalence: 
\begin{assumption}
The trained model $h$ using loss correction (minimizing Eqn. (\ref{eqn:lc2})) is able to memorize $\mathbf{y}_{\text{\emph{LC}}}$. 
\label{ass:lc:memorize}
\end{assumption}
\vspace{-0.05in}
The first message we are ready to send is: \textbf{ For an $x$ with large $l$,  with high probability, loss correction returns smaller generalization error than memorizing noisy labels.} Denote by $sgn(y)$ the sign function of $y$. For $x \in X_{S=l}$, consider a non-trivial case that $\tilde{\p}[\tilde{y} \neq y|x] > 0$\footnote{For trivial cases, our claims would simply be that loss correction performs equally well since the memorizing the noisy label is already equivalent with memorizing the true label.}:
\begin{theorem}
For an $x \in X_{S=l}$ with true label $y$, w.p. at least $1-e^{-2l (1/2-e_{sgn(y)}(x))^2}$, $h$ memorizing $\mathbf{y}_{\text{\emph{LC}}}$ returns a lower error $\p[h(x) \neq y]$ than memorizing the noisy label s.t. $\p[h(x) = k]~ =~ \tilde{\p}[\tilde{y}=k|x]$.  
\label{thm:losscorrection}
\end{theorem}
\vspace{-0.05in}
The above theorem implies when $l \geq \frac{\log 1/\delta}{2\left(\frac{1}{2}-e_+(x)\right)^2}$, memorizing $\smy$ improves the excessive generalization error with probability at least $1-\delta$.
As a corollary of Theorem \ref{thm:losscorrection}: 
\begin{corollary}
For an $x \in X_{S=l}$  with true label $y$, w.p. at least $1-e^{-2l \left(\frac{1}{2}-e_{sgn(y)}(x)\right)^2}$, performing loss correction for $x \in X_{S=l}$ improves the excessive generalization error $\err^{+}_l(\Pd,\A, x|S)$ by 
\vspace{-0.05in}
\[
\Omega\left( \frac{l^2}{n^2} \cdot \textsf{weight}\left(\pi,\left[\frac{2}{3}\frac{l-1}{n-1}, \frac{4}{3}\frac{l}{n}\right] \right)\right)
\]
\label{cor:losscorrection}
\end{corollary}
\vspace{-0.2in}
The above corollary is easily true due to definition of $\err^{+}_l(\Pd,\A, x|S)$, Theorem \ref{thm:noisy}
 \& \ref{thm:losscorrection}, as well as Assumption \ref{ass:lc:memorize}.
 
Our next message is: \textbf{For an $x$ with small $l$, loss correction fails with a substantial probability.} 
Denote by $D_{\text{KL}}(\frac{1}{2}\| e)$ the Kullback-Leibler distance between two Bernoulli 0/1 random variables of parameter $1/2$ and $e$. Consider a non-trivial case that $\tilde{\p}[\tilde{y} \neq y|x] < 1$, we prove:
\begin{theorem}
For an $x \in X_{S=l}$ with true label $y$, w.p. at least $\frac{1}{\sqrt{2l}} \cdot e^{-l \cdot D_{\text{\emph{KL}}}\left(\frac{1}{2}\|e_{sgn(y)}(x)\right)}$, $h$ memorizing $\mathbf{y}_{\text{\emph{LC}}}$ returns a higher error $\p[h(x) \neq y]$ than memorizing noisy label $\ny$. 
\label{thm:losscorrection:lower}
\end{theorem}
When $l$ is small, the reported probability in Theorem \ref{thm:losscorrection:lower} is a non-trivial one. Particularly, when $l \leq \frac{\log \frac{1}{\sqrt{2}\delta}}{D_{\text{KL}}(\frac{1}{2}\| e_{sgn(y)}(x))}$, with probability at least $\delta$, memorizing $\smy$ (or performing loss correction) leads to worse generalization power. 


\subsection{Label smoothing}\label{sec:ls}
Denote by $y_{\text{LS}}$ the ``soft label" for the distribution vector $\sy$, and again we call $h$ memorizing $\sy$ if $\p[h(x)=k]=\tilde{\p}[y_{\text{LS}}=k|x], \forall k \in Y$ -  $\tilde{\p}[y_{\text{LS}}=k|x]$ denotes the empirical distribution for $y_{\text{LS}}$ (empirical average of the soft label $y_{\text{LS}}$):
\begin{align}
\tilde{\p}[y_{\text{LS}}= k|x] =  (1-a) \cdot \tilde{\p}[\tilde{y}=k|x] + \frac{a}{m}~.
\end{align} 
Consider the binary classification case.
Denote the following event:
$
    \EE_+:=\{\tilde{\p}[\tilde{y}=+1|x] > \tilde{\p}[\tilde{y}=-1|x]\}
$ and
 $\bar{\EE}_+$ denotes the opposite event $\tilde{\p}[\tilde{y}=+1|x] < \tilde{\p}[\tilde{y}=-1|x]$. For a non-trivial case  $\tilde{\p}[\tilde{y} \neq y|x] \in (0,1)$:
\begin{theorem}
For an $x \in X_{S=l}$ with true label $y=+1$, when $\EE_+$ happens, $h$ memorizing the smooth label $\mathbf{y}_{\text{\emph LS}}$ leads to a higher error $\p[h(x) \neq y]$ than memorizing corrected label $\mathbf{y}_{\text{\emph LC}}$. When $\bar{\EE}_+$ happens, $h$ memorizing $\mathbf{y}_{\text{\emph LS}}$ has a lower error $\p[h(x) \neq y]$ than memorizing $\mathbf{y}_{\text{\emph LC}}$.
\label{thm:labelsmoothing}
\end{theorem}
\vspace{-0.05in}
Similar result can be proved for the case with $y=-1$ but we will not repeat the details.
Repeating the proofs for Theorem \ref{thm:losscorrection:lower}, we can similarly show that when $l$ is small, there is a substantial probability that $\bar{\EE}_+$ will happen $\left(\geq \frac{1}{\sqrt{2l}} \cdot e^{-l \cdot D_{\text{KL}}\left(\frac{1}{2}\| e_{sgn(y)}(x)\right)}\right)$, therefore label smoothing returns a better generalization power than loss correction. Intuitively, consider the extreme case with $l=1$, and label smoothing has a certain correction power even when this single noisy label is wrong. On the other hand, loss correction (and $\smy$) would memorize this single noisy label for $x$. We view label smoothing as a safe way to perform label correction when $l$ is small, and when the noise rate is excessively high such that $\p[\bar{\EE}_+] > \p[\EE_+]$.

\subsection{Peer loss}\label{sec:pl}
\begin{figure}[!ht]
\centering
\includegraphics[width=0.23\textwidth]{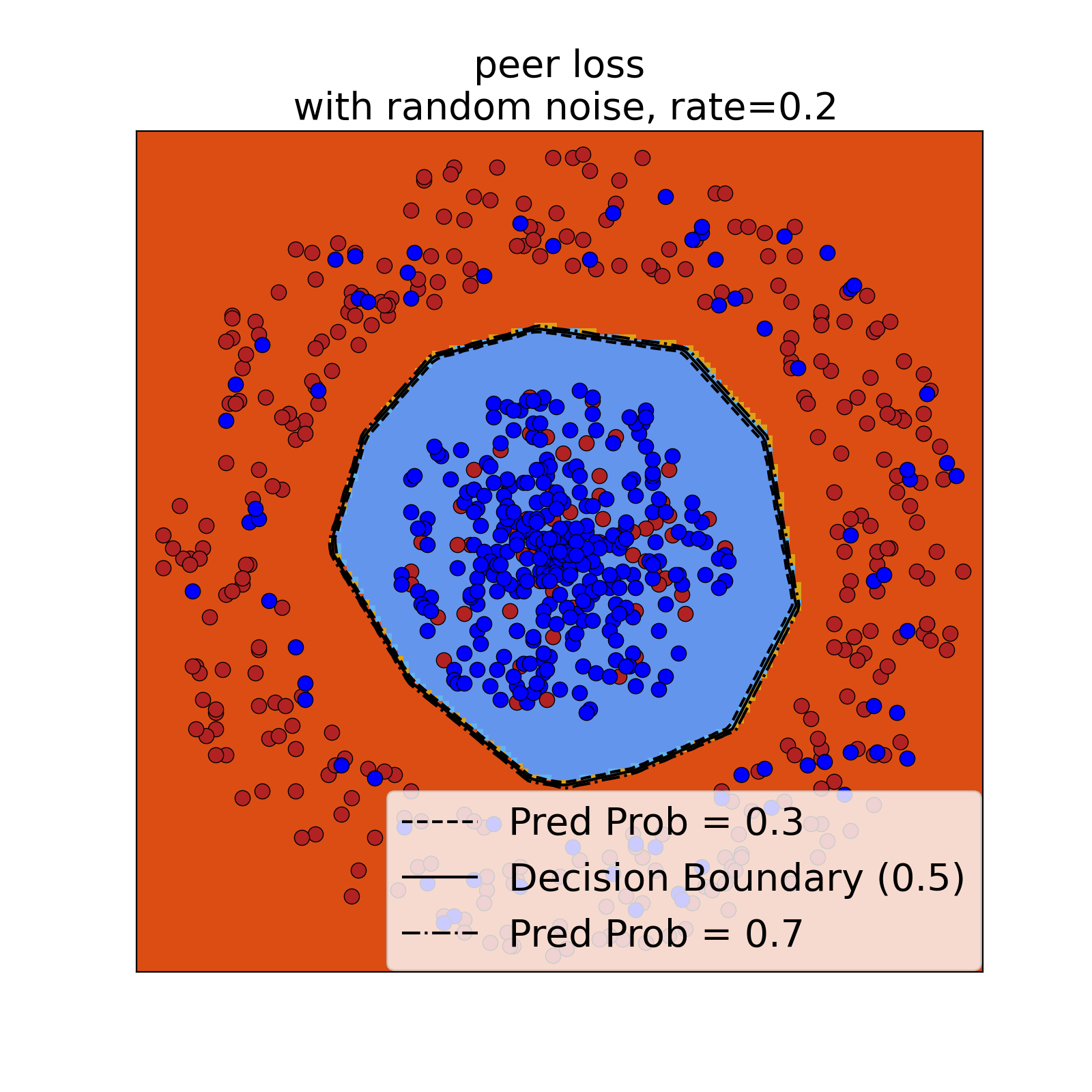}\hspace{-0.2in}
\includegraphics[width=0.23\textwidth]{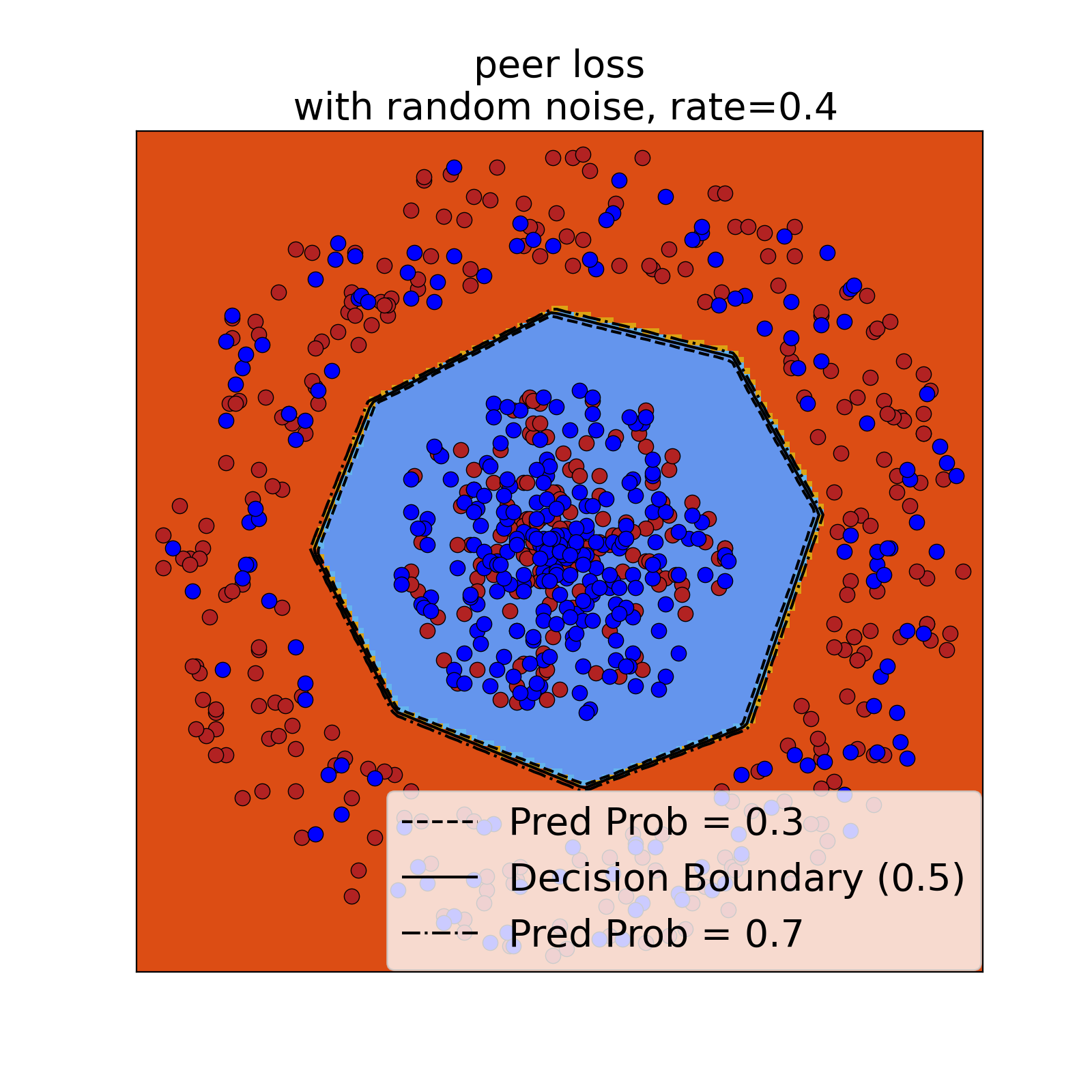}
\vspace{-0.2in}
\caption{2D example with peer loss: $20\%, 40\%$ random noise.}
\label{fig:peerloss}
\end{figure}

The first message we send for peer loss is that: \textbf{For an $x$ with larger $l$, peer loss extremizes $h$'s prediction to the correct label with high probability.}
We first show that peer loss explicitly regularizes $h$ from memorizing noisy labels. We use the cross-entropy loss for $\ell$ in $\ell_{\text{PL}}$ (Eqn. (\ref{eqn:peerloss})). 
\begin{lemma}Denote by $\bQ(x,\tilde{y})$ the joint distribution of $h(x)$ and $\tilde{y}$, $\bP(x),\bP(\tilde{y})$ the marginals of $x,\tilde{y}$, taking expectation of $\ell_{\text{PL}}$ over the training data distribution $ \bP(x,\tilde{y})$, one finds:
\vspace{-0.05in}
\begin{align}
\mathbb{E}_{\mathcal P}\big[\ell_{\text{\emph{PL}}}(h(x), &\tilde{y})\big] = D_{\text{\emph{KL}}}(\bQ(x,\tilde{y})\|\bP(x,\tilde{y}))\nonumber \\
&- D_{\text{\emph{KL}}}(\bQ(x,\tilde{y})\|\bP(x)\times\bP(\tilde{y})). \end{align}
\label{lemma:peerloss}
\end{lemma}
\vspace{-0.25in}
In above we use the standard notation $D_{\text{KL}}$ for KL-divergence between two distributions. While minimizing $D_{\text{KL}}(\bQ(x,\tilde{y})\|\bP(x,\tilde{y})) $ encourages $h$ to reproduce $\bP(x,\tilde{y})$ (the noisy distribution), the second term $D_{\text{KL}}(\bQ(x,\tilde{y})\|\bP(x)\times\bP(\tilde{y}))$ discourages $h$ from doing so by incentivizing $h$ to predict a distribution that is independent from $\tilde{y}$! This regularization power helps lead the training to generate more confident predictions, per a recent result:
\begin{theorem}\label{pro:dynamic}[\cite{cheng2020learning}]
When minimizing $\mathbb{E}\left[\ell_{\text{\emph{PL}}}(h(x), \tilde{y})\right]$, 
solutions satisfying $\p[h(x)=k]>0, \forall k \in Y$ are not optimal. %
\end{theorem}
\vspace{-0.1in}
In the case of binary classification, the above theorem implies that we must have either $\p[h(x)=+1] \rightarrow 1$ or $\p[h(x)=-1] \rightarrow 1$. 
We provide an illustration of this effect in Figure \ref{fig:peerloss}. In sharp contrast to Figure \ref{fig:2d:ce}, the decision boundaries returned by training using peer loss remain tight, despite high presence of label noise. 
\vspace{-0.1in}

\vspace{-0.05in}
Now the question left is: is pushing to confident prediction in the right direction of correcting label noise? If yes, peer loss seems to be achieving the same effect as the loss correction approach, without knowing or using $T(x)$. Consider the binary classification case. 
To simplify our analysis, let's consider only a class-dependent noise setting that $e_+(x) \equiv  e_+, e_-(x) \equiv e_-$. This simplification is needed due to the fact that the definition of peer loss requires drawing a global and random ``peer sample". Technically we can revise peer loss to use ``peer samples" that are similar enough so that they will likely to have the same $e_+(x),e_-(x)$. Denote the priors of the entire distribution by $p_+:=\p_{y' \in \F|S}[y'=+1]>0, p_-:=\p_{y' \in \F|S}[y'=-1]>0$. When $n$ is sufficiently large, we prove: 
\begin{theorem}
For an $x \in X_{S=l}$ with true label $y$, w.p. at least $1-e^{ \frac{-2l}{p^2_{sgn(-y)}\cdot \left(1-e_+-e_-\right)^2}}$, predicting $\p[h(x)=y]=1$ leads to smaller training loss in $\ell_{\text{\emph{PL}}}$ (with $\ell$=CE loss). 
\label{thm:peerloss}
\end{theorem}
\vspace{-0.05in}
Using above theorem and Theorem \ref{thm:noisy}, consider a non-trivial case $\tilde{\p}[\tilde{y} \neq y|x] > 0$ we have:
\begin{corollary}
For an $x \in X_{S=l}$ with true label $y$, w.p. at least $1-e^{ \frac{-2l}{p^2_{sgn(-y)} \cdot \left(1-e_+-e_-\right)^2}}$, 
training using $\ell_{\text{\emph{PL}}}$  improves the individual excessive generalization error  $\err^{+}_l(\Pd,\A, x|S)$ by:
\vspace{-0.1in}
\[
\Omega\left( \frac{l^2}{n^2} \textsf{weight}\left(\pi,\left[\frac{2}{3}\frac{l-1}{n-1}, \frac{4}{3}\frac{l}{n}\right] \right)\cdot \sum_{k \neq y} \tilde{\p}[\tilde{y}=k|x]\right)
\]
\label{cor:peerloss}
\end{corollary}
\vspace{-0.15in}
Both peer loss and loss correction implicitly use the posterior distribution of noisy labels to hopefully extremize $h(x)$ to the correct direction. The difference is peer loss extremizes even more ($\p[h(x)=k] \rightarrow 1$ for some $k$) when it is confident. Therefore, when there is sufficient information (i.e., $l$ being large), peer loss tends to perform better than loss correction which needs explicit knowledge of the true transition matrix and performs a precise and exact bias correction step. This is also noted empirically in \cite{liu2019peer}, and our results provide the theoretical justifications.

Similar to Theorem \ref{thm:losscorrection:lower}, when $l$ is small, the power of peer loss does seem to drop:
 \textbf{For an $x$ with small $l$, peer loss extremizes $h$'s prediction to the wrong label with a substantial probability.} Formally,  
\begin{theorem}
For an $x \in X_{S=l}$ with true label $y$, w.p. at least $ \frac{1}{\sqrt{2l}}e^{-l \cdot D_{\text{\emph{KL}}}\left(\frac{1}{2}\| e_{sgn(y)}\right)}$, predicting $\p[h(x)=-y] = 1$ leads to smaller training loss in $\ell_{\text{\emph{PL}}}$. 
\label{thm:peerloss:lower}
\end{theorem}
\vspace{-0.05in}
For a non-trivial case that $\tilde{\p}[\tilde{y} \neq y|x] < 1$, this implies higher prediction error than memorizing the noisy labels.

\section{Takeaways and Conclusion}

  We studied the impact of a model memorizing noisy labels. 
This paper proved the disparate impact of noisy labels at the instance level, and then the fact that existing treatments can often lead to disparate outcomes, with low-frequency instances being more likely to be mistreated. This observation is particularly concerning when a societal application is considered, and the low-frequency examples are drawn from a historically disadvantaged population (thus low presence in data). Specifically:
    
\textbf{Frequent instance} While high-frequent instances (large $l$) have a higher impact on the generalization bound, and miss-classifying one such example would be more costly, our analysis shows that due to the existence of a good number of noisy labels, existing approaches can often counter the negative effects with high probability. 

\textbf{Long-tail instance} For a rare instance $x$ with small $l$, while missing it would incur a much smaller penalty in generalization, still, its impact is non-negligible. Moreover, due to the severely limited label information, we find the noise correction approaches would have a substantial probability of failing to correct such cases. 

The above observations require us to rethink the distribution of label noise across instances and might potentially require different treatments for instances in different regimes.  

\textbf{Acknowledgments} This work is partially supported by the National Science Foundation (NSF) under grant IIS-2007951 and the NSF FAI program in collaboration with Amazon under grant IIS-2040800. The author thanks Zhaowei Zhu, Xingyu Li and Jiaheng Wei for helps with Figure \ref{fig:cifar}~-~\ref{fig:peerloss}, as well as Lemma \ref{lemma:peerloss}. The author thanks anonymous ICML reviewers for their comments that improved the presentation of the paper. The final draft also benefited substantially from discussions with Gang Niu.  




\nocite{langley00}
\bibliographystyle{icml2021}

\bibliography{references, noise_learning, library,myref}

\appendix
\newpage
\onecolumn
\appendix

\begin{center}
    \section*{\Large Appendix}
\end{center}
The Appendix is organized as follows. 
\squishlist
\item Section A presents omitted proofs for theoretical conclusions in the main paper. 
\item Section B includes additional details for Figures \ref{fig:2d:ce} and \ref{fig:cifar-memorization}.
\item Section C provides additional discussions. 
\squishend

\section{Proofs}
\section*{Proof for Theorem \ref{thm:main:tau}}

\begin{proof}
Recall the definition of $\tau_l$:
\begin{align}
    \tau_l := \frac{\E_{\alpha \sim \bar{\pi}^N}[\alpha^{l+1}\cdot (1-\alpha)^{n-l}]}{\E_{\alpha \sim \bar{\pi}^N}[\alpha^{l}\cdot (1-\alpha)^{n-l}]}
\end{align}
To bound $\tau_l$, we start with the denominator $\E_{\alpha \sim \bar{\pi}^N}[\alpha^{l}\cdot (1-\alpha)^{n-l}].$ First $\forall \alpha \in [0,1]$, we have 
 \[
 \alpha^{l-1}\cdot (1-\alpha)^{n-l} \leq \left(\frac{l-1}{n-1}\right)^{l-1} \left(1-\frac{l-1}{n-1}\right)^{n-l}~.
 \]
 The above can be easily verified by the checking the first order condition of 
$
 \log \left( \alpha^{l-1}\cdot (1-\alpha)^{n-l}\right).
$ Therefore 
 \begin{align}
     &\E_{\alpha \sim \bar{\pi}^N}[\alpha^{l}\cdot (1-\alpha)^{n-l}]\nonumber \\
     \leq &\left(\frac{l-1}{n-1}\right)^{l-1} \left(1-\frac{l-1}{n-1}\right)^{n-l} \cdot \E_{\alpha \sim \bar{\pi}^N}[\alpha]\nonumber \\
     = &\left(\frac{l-1}{n-1}\right)^{l-1} \left(1-\frac{l-1}{n-1}\right)^{n-l} \cdot \frac{1}{N}~.\label{eqn:deno}
 \end{align}
 where the last equality is due to $\bar{\pi}^N[\alpha] = \frac{1}{N}$. Now let's look at the numerator $\E_{\alpha \sim \bar{\pi}^N}[\alpha^{l+1}\cdot (1-\alpha)^{n-l}]$. For $\alpha \in [\frac{l-1}{n-1}, \frac{l}{n}]$, we have
 \begin{align*}
     & \alpha^{l+1}\cdot (1-\alpha)^{n-l} \geq \alpha \cdot \left( \frac{l-1}{n-1} \right)^{l} \cdot \left(1-\frac{l}{n}\right)^{n-l} 
 \end{align*}
 Therefore combining the bounds for both denominator and the numerator we have
 \begin{align}
     \tau_l \geq & \frac{ \left( \frac{l-1}{n-1} \right)^{l} \cdot \left(1-\frac{l}{n}\right)^{n-l} }{\left(\frac{l-1}{n-1}\right)^{l-1} \left(1-\frac{l-1}{n-1}\right)^{n-l}} \cdot \E\left [N \alpha\cdot \mathds{1}\left(\alpha \in \left[\frac{l-1}{n-1},\frac{l}{n}\right]\right)\right]\nonumber \\
     =& \underbrace{\frac{l-1}{n-1}}_{\text{Term 1}} \cdot \underbrace{\left(\frac{1-\frac{l}{n}}{1-\frac{l-1}{n-1}}\right)^{n-l}}_{\text{Term 2}} \cdot \underbrace{\textsf{weight}\left(\bar{\pi}^N,\left[\frac{l-1}{n-1},\frac{l}{n}\right]\right)}_{\text{Term 3}},\label{eqn:tau:main}
 \end{align}
 where the $\textsf{weight}$ is introduced by its definition. 
 For the second term above,
 \begin{align}
 &\left(\frac{1-\frac{l}{n}}{1-\frac{l-1}{n-1}}\right)^{n-l}\nonumber  \\
 =&\left(1- \frac{\frac{l}{n}-\frac{l-1}{n-1}}{1-\frac{l-1}{n-1}}\right)^{n-l}\nonumber \\
 =&\left(1- \frac{l(n-1)-(l-1)n}{n(n-1)-(l-1)n}\right)^{n-l}\nonumber \\
 =&\left(1- \frac{n-l}{n(n-l)}\right)^{n-l}\nonumber \\
 \geq & 1- \frac{n-l}{n(n-l)} \cdot (n-l) =\frac{l}{n}.\label{eqn:tau:term2}
 \end{align}
 For the third term, we will call Lemma 2.4 of \citet{feldman2020does}: 
 \begin{lemma}[Lemma 2.4 of \cite{feldman2020does}] For any $0<\beta_1 < \beta_2 < 1$, and for any $\gamma>0$, 
\begin{align}
     \textsf{weight}\left(\bar{\pi}^N,\bigl[\beta_1,\beta_2\bigr]\right) \geq \frac{1-\delta}{1-\frac{1}{N}+\beta_2+\gamma} \cdot   \textsf{weight}\left(\bar{\pi}^N,\left[\frac{\beta_1}{1-\frac{1}{N}+\beta_1-\gamma},\frac{\beta_2}{1-\frac{1}{N}+\beta_2+\gamma} \right]\right),
\end{align}
 where in above  $\delta:= 2 \cdot e^{\frac{-\gamma^2}{2(N-1) Var(\pi)+2\gamma \pi_{\max}/3}}$, and 
 \begin{align*}
     Var(\pi):= \sum_{j \in [N]} (\pi_j -\frac{1}{N}) \leq \frac{\pi_{\max}}{N}.
 \end{align*}
\label{lemma:bound:weight}
\end{lemma}
Using above, we next further derive that \vspace{-2pt}
\begin{tcolorbox}[colback=grey!5!white,colframe=grey!5!white]
\vspace{-10pt}
  \begin{align*}
     &\textsf{weight}\bigl(\bar{\pi}^N,\bigl[\frac{l-1}{n-1},\frac{l}{n}\bigr]\bigr) \geq  0.4\cdot  \textsf{weight}\left(\pi,\left[\frac{2}{3}\frac{l-1}{n-1}, \frac{4}{3}\frac{l}{n}\right]\right)
 \end{align*}
 \vspace{-18pt}
\end{tcolorbox}
\vspace{-2pt}
To see this, using above Lemma \ref{lemma:bound:weight}, for any $\gamma > 0$,
 \begin{align*}
     &\textsf{weight}\left(\bar{\pi}^N,\left[\frac{l-1}{n-1},\frac{l}{n}\right]\right) \geq  \frac{1-\delta}{1-\frac{1}{N}+\frac{l}{n} + \gamma} \cdot\textsf{weight}\left(\pi,\left[\frac{\frac{l-1}{n-1}}{1-\frac{1}{N}+\frac{l-1}{n-1}-\gamma},\frac{\frac{l}{n}}{1-\frac{1}{N}+\frac{l}{n}+\gamma}\right]\right)
 \end{align*}

 Let $\gamma = \frac{1}{2}$, for sufficiently large $n (\gg l),N$, 
 \[
 1-\frac{1}{N}+\frac{l-1}{n-1}-\gamma \leq \frac{2}{3} 
 \]
 Similarly 
 \[
 1-\frac{1}{N}+\frac{l}{n}+\gamma \geq \frac{4}{3}
 \]
 Easy to show that $\delta \leq 2 e^{-1/10\pi_{max}}$, and when $\pi_{max} \leq \frac{1}{20}$, we have $\delta \leq 2 e^{-2} \leq 0.3$. Therefore 
 \begin{align}
     \frac{1-\delta}{1-\frac{1}{N}+\frac{l}{n} + \gamma} \geq \frac{0.7}{7/4} = 0.4.
 \end{align}
 To summarize 
  \begin{align*}
     &\textsf{weight}\bigl(\bar{\pi}^N,\bigl[\frac{l-1}{n-1},\frac{l}{n}\bigr]\bigr) \geq  0.4\cdot  \textsf{weight}\left(\pi,\left[\frac{2}{3}\frac{l-1}{n-1}, \frac{4}{3}\frac{l}{n}\right]\right)
 \end{align*}
 Combining Term 2 and 3 we complete the proof. 
 \end{proof}
 
\section*{Proof for Proposition \ref{prop:tau}}
\begin{proof}
 The major difference from proving Theorem \ref{thm:main:tau} is due to the reasoning of the numerator: $\E_{\alpha \sim \bar{\pi}^N}[\alpha^{l+1}\cdot (1-\alpha)^{n-l}]$. Now for $\alpha \in [\frac{l-1}{1.1(n-1)}, \frac{l-1}{n-1}]$, we have
 \begin{align*}
     & \alpha^{l+1}\cdot (1-\alpha)^{n-l}  \geq  \alpha \cdot \left( \frac{l-1}{1.1(n-1)} \right)^{l} \cdot \left(1-\frac{l-1}{n-1}\right)^{n-l} 
 \end{align*}
 Therefore, using Eqn. (\ref{eqn:deno}), and the definition of $\tau_l$ we have
 \begin{align}
     \tau_l \geq & \frac{ \left( \frac{l-1}{1.1(n-1)} \right)^{l} \cdot \left(1-\frac{l-1}{n-1}\right)^{n-l} }{\left(\frac{l-1}{n-1}\right)^{l-1} \left(1-\frac{l-1}{n-1}\right)^{n-l}} \cdot \E\left[N \alpha\cdot \mathds{1}\left(\alpha \in \left[\frac{l-1}{1.1 (n-1)},\frac{l-1}{n-1}\right]\right)\right]\nonumber \\
     =& \frac{l-1}{n-1} \cdot \frac{1}{1.1 ^l} \cdot \textsf{weight}\left(\bar{\pi}^N,\left[\frac{l-1}{1.1 (n-1)},\frac{l-1}{n-1}\right]\right).\label{eqn:tau:2}
 \end{align}
Again, using Lemma 2.4 of \citet{feldman2020does}, for the $\textsf{weight}$ term, we further derive that for any $\gamma > 0$,
\vspace{-2pt}
\begin{tcolorbox}[colback=grey!5!white,colframe=grey!5!white]
\vspace{-10pt}
\begin{small}
 \begin{align*}
     &\textsf{weight}\left(\bar{\pi}^N,\left[\frac{l-1}{1.1(n-1)},\frac{l-1}{n-1}\right]\right)\geq  \frac{1-\delta}{1-\frac{1}{N}+\frac{l-1}{n-1} + \gamma} \cdot\textsf{weight}\left(\pi,\left[\frac{\frac{l-1}{1.1(n-1)}}{1-\frac{1}{N}+\frac{l-1}{1.1(n-1)}-\gamma},\frac{\frac{l-1}{n-1}}{1-\frac{1}{N}+\frac{l-1}{n-1}+\gamma}\right]\right)
 \end{align*}
\end{small}
 \vspace{-18pt}
\end{tcolorbox}
\vspace{-2pt}
Let $\gamma = \frac{1}{2}$, for sufficiently large $n (\gg l),N$, 
 \[
 1-\frac{1}{N}+\frac{l-1}{1.1 (n-1)}-\gamma \leq \frac{2}{3}
 \]
 Similarly 
 \[
 1-\frac{1}{N}+\frac{l-1}{1.1(n-1)}+\gamma \geq \frac{4}{3}
 \]
  Similarly we show that $\delta \leq 2 e^{-1/10\pi_{max}}$, and when $\pi_{max} \leq \frac{1}{20}$ we have $\delta \leq 2 e^{-2} \leq 0.3$. Therefore 
 \begin{align}
     \frac{1-\delta}{1-\frac{1}{N}+\frac{l-1}{n-1} + \gamma} \geq \frac{0.7}{7/4} = 0.4.
 \end{align}
 To summarize 
  \begin{align*}
     &\textsf{weight}\left(\bar{\pi}^N,\left[\frac{l-1}{1.1(n-1)},\frac{l-1}{n-1}\right]\right) \geq  \textsf{weight}\left(\pi,\left[0.7 \cdot \frac{l-1}{n-1}, \frac{4}{3}\frac{l-1}{n-1}\right]\right)
 \end{align*}
 Putting this above bound back to Eqn. (\ref{eqn:tau:2}) we complete the proof. 
 \end{proof}

 \section*{Proof for Theorem \ref{thm:noisy}}
 
 \begin{proof}
This is simply because when $h$ memorizes the noisy labels for $x$ such that $\p_{h \sim \A(S')}[h(x) = k] = \tilde{\p}[\tilde{y}=k|x]$, we will have:
\begin{align*}
&\err^{+}_l(\Pd,\A, x|S)\\
  =&\tau_l \cdot \p_{h \sim \A(S')}[h(x) \neq y] \\
 =&\tau_l \cdot \sum_{k \neq y}\tilde{\p}[\tilde{y}=k|x].
\end{align*}
Plugging the lower bound we prepared for $\tau_l$ earlier (Theorem \ref{thm:main:tau}) we proved the claim.
\end{proof}

\section*{Proof for Lemma \ref{lemma:smy}}

\begin{proof}
Using the definition of $\smy$, and using the knowledge of Eqn. (\ref{eqn:T:inverse}) we know that
\begin{align*}
    &\smy[1] = \frac{1}{1-e_+(x)-e_-(x)}\left((1-e_+(x)) \tilde{\p}[\tilde{y}=-1|x] - e_+(x) \tilde{\p}[\tilde{y}=+1|x]  \right)\\
    &\smy[2] = \frac{1}{1-e_+(x)-e_-(x)}\left((1-e_-(x)) \tilde{\p}[\tilde{y}=+1|x] - e_-(x) \tilde{\p}[\tilde{y}=-1|x]  \right)\\
\end{align*}
Therefore 
\begin{align*}
    \smy[1]+\smy[2] = \frac{1}{1-e_+(x)-e_-(x)}\left((1-e_+(x)-e_-(x))\cdot \tilde{\p}[\tilde{y}=-1|x]+(1-e_+(x)-e_-(x))\cdot \tilde{\p}[\tilde{y}=+1|x] \right) = 1.
\end{align*}
\end{proof}

 \section*{Proof for Theorem \ref{thm:losscorrection}}

\begin{proof}

 Due to symmetricity, we consider $y=+1$.
Using the definition of $\smy$, and using the knowledge of Eqn. (\ref{eqn:T:inverse}) we know that
\begin{align*}
    \p[y_{\text{LC}} = +1|x] = \frac{(1-e_-(x)) \cdot \tilde{\p}[\tilde{y}=+1|x]-e_+(x)\cdot \tilde{\p}[\tilde{y}=-1|x]}{1-e_+(x)-e_-(x)}
\end{align*}
Recall we denote by $y_{\text{LC}}$ the random variable drawn according to $\smy$.
Next we show:
\begin{align}
\tilde{\p}[\tilde{y}=+1|x] > \tilde{\p}[\tilde{y}=-1|x] \Leftrightarrow \p[y_{\text{LC}} = +1|x] > \tilde{\p}[\tilde{y}=+1|x]. \label{eqn:equiv}
\end{align}
This is equivalent to the following comparison (when $e_+(x)+e_-(x) < 1$):
\begin{align*}
&\text{RHS of Eqn. (\ref{eqn:equiv})}\\
\Leftrightarrow ~ &\frac{(1-e_-(x)) \cdot \tilde{\p}[\tilde{y}=+1|x]-e_+(x) \cdot \tilde{\p}[\tilde{y}=-1|x]}{1-e_+(x)-e_-(x)} >  \tilde{\p}[\tilde{y}=+1|x]\\
  \Leftrightarrow~ & (1-e_-(x))\cdot \tilde{\p}[\tilde{y}=+1|x]-e_+(x) \cdot \tilde{\p}[\tilde{y}=-1|x] > (1-e_+(x)-e_-(x)) \cdot \tilde{\p}[\tilde{y}=+1|x]\\
  \Leftrightarrow ~& e_+(x) \cdot \tilde{\p}[\tilde{y}=+1|x] >e_+\cdot\tilde{\p}[\tilde{y}=-1|x]\\
  \Leftrightarrow ~& \tilde{\p}[\tilde{y}=+1|x] > \tilde{\p}[\tilde{y}=-1|x]\\
  \Leftrightarrow ~&\text{LHS of Eqn. (\ref{eqn:equiv})}
\end{align*}
Note that because we consider a non-trivial case $\tilde{\p}[\tilde{y} \neq y|x] > 0 $, we have  $\tilde{\p}[\tilde{y}=+1|x] < 1$. Therefore the above derivation holds even if we capped $ \p[y_{\text{LC}} = +1|x]=\frac{(1-e_-(x)) \cdot \tilde{\p}[\tilde{y}=+1|x]-e_+(x) \cdot \tilde{\p}[\tilde{y}=-1|x]}{1-e_+(x)-e_-(x)}$ at 1 to make it a valid probability measure. 

Now we derive when $\tilde{\p}[\tilde{y}=+1|x] > \tilde{\p}[\tilde{y}=-1|x]$. Recall $\tilde{y}(1),...,\tilde{y}(l)$ denote the $l$ noisy labels for $x \in X_{S=l}$, and let $Z_1,...,Z_l$ denote the $l$ Bernoulli random variable that $Z_k = \mathds{1}(\tilde{y}(k) = +1), k \in [l]$. Then 
\begin{align*}
& \tilde{\p}[\tilde{y}=+1|x] > \tilde{\p}[\tilde{y}=-1|x]\\
\Leftrightarrow ~&  \tilde{\p}[\tilde{y}=+1|x] > 1/2\\
\Leftrightarrow ~&
\frac{\sum_{k \in [l]}Z_k}{l} > 1/2.
\end{align*}
Now we derive $\p\left[\frac{\sum_{k \in [l]}Z_k}{l} > 1/2\right]
$. 
Note $\E\left[\frac{\sum_{k \in [l]}Z_k}{l}\right] = 1-e_+(x)$. By applying Hoeffding inequality we prove that
\[
\p\left[\frac{\sum_{k \in [l]}Z_k}{l} \leq 1/2\right] \leq e^{-2l \left(1/2-e_+(x)\right)^2}.
\]
Therefore 
\[
\p\left[\frac{\sum_{k \in [l]}Z_k}{l} > 1/2\right]\geq 1-e^{-2l \left(1/2-e_+(x)\right)^2}.
\]
The case with $y=-1$ is entirely symmetric so we omit the details.
\end{proof}

\section*{Proof for Corollary \ref{cor:losscorrection}}
\begin{proof}
Because 
$
\err^{+}_l(\Pd,\A, x|S):= \tau_l \cdot \p_{h \sim \A(S')}[h(x) \neq y]
$, as well as w.p. at least $1-e^{-2l (1/2-e_{sgn(y)}(x))^2}$, memorizing $\smy$ returns lower error $\p[h(x) \neq y]$ than memorizing the noisy labels s.t. $\p[h(x) = k] = \tilde{\p}[\tilde{y}=k|x]$, the corollary is then true via the lower bound for $\tau_l$. 
\end{proof}

 \section*{Proof for Theorem \ref{thm:losscorrection:lower}}

\begin{proof}
Again due to symmetricity, we consider $y=+1$.
Similar as argued in the proof of Theorem \ref{thm:losscorrection}, from Eqn. (\ref{eqn:equiv}) we know that when $\tilde{\p}[\tilde{y}=+1|x] < \tilde{\p}[\tilde{y}=-1|x]$, we have $ \p[ y_{\text{LC}}= +1|x] < \tilde{\p}[\tilde{y}=+1|x]$, that is memorizing $\smy$ is worse than memorizing $\ny$. Again because we consider a non-trivial case $\tilde{\p}[\tilde{y} \neq y|x] < 1$, we have  $\tilde{\p}[\tilde{y}=+1|x] > 0$. Therefore the above derivation holds even if we capped $ \p[y_{\text{LC}} = +1|x]=\frac{(1-e_-(x)) \cdot \tilde{\p}[\tilde{y}=+1|x]-e_+(x) \cdot \tilde{\p}[\tilde{y}=-1|x]}{1-e_+(x)-e_-(x)}$ at 0 to make it a valid probability measure.

Similarly define $Z_k$ for $k \in [l]$ as in Proof for Theorem \ref{thm:losscorrection}: the $l$ Bernoulli random variable that $Z_k = \mathds{1}(\tilde{y}(k) = +1), k \in [l]$. Then  
\[
\p[y_{\text{LC}}= +1|x] < \tilde{\p}[\tilde{y}=+1|x] \Leftrightarrow \tilde{\p}[\tilde{y}=+1|x] < \tilde{\p}[\tilde{y}=-1|x] \Leftrightarrow \frac{\sum_{k \in [l]}Z_k}{l} < 1/2
\]
We bound $\p\left[\frac{\sum_{k \in [l]}Z_k}{l} < 1/2\right]$:
\begin{align*}
&\p\left[\frac{\sum_{k \in [l]}Z_k}{l} < 1/2\right]\\
=&\p\left[\sum_{k \in [l]}1-Z_k \geq l/2\right]\\
=&\p[\text{Bin}(l,e_+(x)) \geq l/2],
\end{align*}
where we use $\text{Bin}$ to denote a Binomial random variable, and the fact that $\E\left[\frac{\sum_{k \in [l]}(1-Z_k)}{l}\right] = e_+(x)$. Using tail bound for Bin (e.g., Lemma 4.7.2 of \cite{ash1990}) yields:
\begin{align*}
    \p[\text{Bin}(l,e_+(x)) \geq l/2] \geq& \frac{1}{\sqrt{8 \cdot \frac{1}{2}l (1-\frac{1}{2})}} \cdot e^{-l \cdot D_{\text{KL}}\left(\frac{1}{2}\| e_+(x)\right)}\\
    =& \frac{1}{\sqrt{2l}} \cdot e^{-l \cdot D_{\text{KL}}\left(\frac{1}{2}\| e_+(x)\right)},
\end{align*}
completing the proof. The case with $y=-1$ is entirely symmetric so we omit the details.
\end{proof}

 \section*{Proof for Theorem \ref{thm:labelsmoothing}}
 
 \begin{proof}
Note that label smoothing smooths the noisy labels in the following way:
\begin{align}
\tilde{\p}[y_{\text{LS}}= +1|x] =  (1-a) \cdot \tilde{\p}[\tilde{y}=+1|x] + \frac{a}{2}.~
\end{align} 
The proof is then simple: when $\EE_+$ is true, from the proof of Theorem \ref{thm:losscorrection}, Eqn. (\ref{eqn:equiv}),  we know that $\smy$ further extremizes/increases the prediction of $+1$ that $\p[y_{\text{LC}}= +1|x]>\tilde{\p}[\tilde{y}=+1|x]$, while label smoothing $y_{\text{LS}}$ reduces from $\tilde{\p}[\tilde{y}=+1|x] $ by a factor of $a \cdot \tilde{\p}[\tilde{y}=+1|x] - \frac{a}{2}>0 $. Therefore, memorizing smoothed label increases the error $\p[h(x) \neq y]$. 

The above observation is reversed when the opposite event $\bar{\EE}_+$ is instead true: in this case, $\smy$ further extremizes/increases the prediction of $-1$ that $\p[y_{\text{LC}}= +1|x]<\tilde{\p}[\tilde{y}=+1|x]$,  while label smoothing $y_{\text{LS}}$ increases from $\tilde{\p}[\tilde{y}=+1|x] $ by a factor of $\frac{a}{2} - a \cdot \tilde{\p}[\tilde{y}=+1|x] >0 $.
\end{proof}
 
 \section*{Proof for Lemma \ref{lemma:peerloss}}
 
 \begin{proof}
Denote by $\ell_{\text{CE}}$ the CE loss, and $\calD_{x},\calD_{\tilde{y}}$ the marginal distribution of $x,\tilde{y}$ explicitly. 
\begin{equation}\label{Epeer0}
	\begin{split}
&\mathbb{E}_{x \times \tilde{y}}\left[\ell_{\text{PL}}(h(x), \tilde{y})\right] 
\\
=&\mathbb{E}_{x \times \tilde{y}}\left[\ell_{\text{CE}}(h(x), \tilde{y})\right]-\mathbb{E}_{\calD_{\tilde{y}}}\left[\mathbb{E}_{\calD_{x}}\left[\ell_{\text{CE}}\left(h\left(x\right), \tilde{y}\right)\right]\right]\\
=& \underbrace{-\sum_{x \in X}\sum_{\tilde{y}\in Y} \p(x,\tilde{y})\log{\bQ(\tilde{y}|x)}}_{\text{CE term}}  +\underbrace{\sum_{x \in X} \sum_{\tilde{y}\in Y} \bP(x)\bP(\tilde{y})\log{\bQ(\tilde{y}|x)}}_{\text{Peer term}},
	\end{split}
\end{equation}
The conditional probability $\bQ(\tilde{y}|x)$ is defined as the prediction of the underlying neural network model, while $\bP(x)\bP(y)$ captures the probabilities of the marginal-product of the training distribution. We call the first sum-integration as {\it CE term} and the second as {\it peer term}.

 For the CE term, we further have
\begin{align*}
&-\sum_{x \in X}\sum_{\tilde{y}\in Y} \p(x,\tilde{y})\log{\bQ(\tilde{y}|x)} \\
=& -\sum_{x \in X}\sum_{\tilde{y}\in Y} \left[ \bP(x,\tilde{y})\log{\bQ(\tilde{y}|x)\bP(x)} -\bP(x,\tilde{y})\log\bP(x,\tilde{y}) \right]\\
	=& -\sum_{x \in X}\sum_{\tilde{y}\in Y}  \bP(x,\tilde{y})\log\frac{\bQ(x,\tilde{y})}{\bP(x,\tilde{y})}\\
	=& D_{\text{KL}}(\bQ(x,\tilde{y})\|\bP(x,\tilde{y}))
\end{align*}
In the above, we used $\bQ(x,\tilde{y}) = \bQ(\tilde{y}|x)\bP(x)$ as in classification task the model prediction does not affect the feature distribution. Similarly, for the peer term
\begin{align*}
&\sum_{x \in X} \sum_{\tilde{y}\in Y} \bP(x)\bP(\tilde{y})\log{\bQ(\tilde{y}|x)}\\
=& \sum_{x \in X} \sum_{\tilde{y}\in Y}  \biggl[ \bP(x)\bP(\tilde{y})\log{\bQ(\tilde{y}|x)\bP(x)} - \bP(x)\bP(\tilde{y})\log\bP(x)\bP(\tilde{y})\biggr]\\
	=& \sum_{x \in X} \sum_{\tilde{y}\in Y} \bP(x)\bP(\tilde{y})\log\frac{\bQ(x,\tilde{y})}{\bP(x)\bP(\tilde{y})}\\
	=& -D_{\text{KL}}(\bQ(x,\tilde{y})\|\bP(x)\times\bP(\tilde{y})).
\end{align*}
Combing the above derivations for the CE and peer term we complete the proof.
\end{proof}

\section*{Proof for Theorem \ref{thm:peerloss}}

\begin{proof}
Again due to symmetricity, we consider $y=+1$.
It was shown in \cite{cheng2020learning} that, when taking expectation over the data distribution, the original definition of peer loss 
\begin{align}
    \ell_{\text{PL}}(h(x),\tilde{y}) := \ell(h(x),\tilde{y}) - \ell(h(x_{p_1}),\tilde{y}_{p_2})
    \end{align}
is equivalent to
\begin{align}
\ell(h(x), \tilde{y}) - \tilde{\mathbb E}[\ell(h(x),\tilde{y}_{q})],
    \end{align}
where $q$ is a randomly sampled index from $[n]$, and the expectation is over the randomness in $\tilde{y}_{q}$. The high-level intuition is that each $x$ appears in the peer terms exactly once in expectation, so we can fix $x_{p_1}$ be the same $x$. Then we will only vary $p_2$ (index $q$ in our notation). The operation of taking expectation is to reduce uncertainty in the peer term.

The average empirical peer loss on $x$ is then given by
\begin{align*}
    &\frac{1}{l}\sum_{i=1}^l \ell_{\text{PL}}(h(x),\tilde{y}(i))\\
    = &\frac{1}{l}\sum_{i=1}^l \ell(h(x), \tilde{y}) - \tilde{\mathbb E}[\ell(h(x),\tilde{y}_{q})]\\
    =&\tilde{\mathbb E}\left[\ell(h(x), \tilde{y})\right] -\ell(h(x),-1) \cdot \tilde{\p}[\tilde{y}_{q}=-1]\\
    &~~~~- \ell(h(x),+1) \cdot  \tilde{\p}[\tilde{y}_{q}=+1].
\end{align*}
where $\tilde{\mathbb E}$ denotes the empirical expectation w.r.t. the empirical distribution of $\tilde{y}|x$, and $\tilde{\p}[\tilde{y}_{q}]$s are the empirical distribution of $\tilde{y}_q$.
As shown already, peer loss pushes $h$ to predict one class confidently, therefore there are two cases: $h(x)=+1$ or $h(x)=-1$. 

When $h(x)=+1$ we have 
\begin{align*}
    &\tilde{\mathbb E}\left[\ell(h(x)=+1, \tilde{y})\right] -\ell(h(x),-1) \cdot \tilde{\p}[\tilde{y}_{q}=-1]\\
    &~~~~- \ell(h(x),+1) \cdot  \tilde{\p}[\tilde{y}_{q}=+1]\\
    =&\tilde{\p}[\tilde{y}=+1|x] \cdot \ell(h(x)=+1,+1) +\tilde{\p}[\tilde{y}=-1|x] \cdot \ell(h(x)=+1,-1)\\
    &~~~~-\ell(h(x)=+1,-1) \cdot \tilde{\p}[\tilde{y}_{q}=-1]- \ell(h(x)=+1,+1) \cdot  \tilde{\p}[\tilde{y}_{q}=+1]\\
    &=\left(\tilde{\p}[\tilde{y}=+1|x] - \tilde{\p}[\tilde{y}_{q}=+1]\right) \cdot \left(\ell(h(x)=+1,+1) -\ell(h(x)=+1,-1)\right)
\end{align*}
While if $h(x)=-1$, we have
\begin{align*}
    &\tilde{\mathbb E}\left[\ell(h(x)=-1, \tilde{y})\right] -\ell(h(x),-1) \cdot \tilde{\p}[\tilde{y}_{q}=-1]\\
    &~~~~- \ell(h(x),+1) \cdot  \tilde{\p}[\tilde{y}_{q}=+1]\\
    =&\tilde{\p}[\tilde{y}=+1|x] \cdot \ell(h(x)=-1,+1) +\tilde{\p}[\tilde{y}=-1|x] \cdot \ell(h(x)=-1,-1)\\
    &~~~~-\ell(h(x)=-1,-1) \cdot \tilde{\p}[\tilde{y}_{q}=-1]- \ell(h(x)=-1,+1) \cdot  \tilde{\p}[\tilde{y}_{q}=+1]\\
    &=\left(\tilde{\p}[\tilde{y}=-1|x] - \tilde{\p}[\tilde{y}_{q}=-1]\right)\left(\ell(h(x)=-1,-1) -\ell(h(x)=-1,+1)\right)\\
    &=-\left(\tilde{\p}[\tilde{y}=+1|x] - \tilde{\p}[\tilde{y}_{q}=+1]\right)\left(\ell(h(x)=-1,-1) -\ell(h(x)=-1,+1)\right)
\end{align*}
For CE, we have $\ell(h(x)=+1,+1) = \ell(h(x)=-1,-1)=0$, and in \citet{cheng2020learning,liu2019peer}, $\ell(h(x)=-1,+1) = \ell(h(x)=+1,-1)$ is set to be a large positive quantity. Let's denote it as $C$. Then $h(x)=+1$ returns a lower loss ($\left(\tilde{\p}[\tilde{y}=+1|x] - \tilde{\p}[\tilde{y}_{q}=+1] \right) \cdot (-C)$, a negative quantity as compared to $-\left(\tilde{\p}[\tilde{y}=+1|x] - \tilde{\p}[\tilde{y}_{q}=+1]\right ) \cdot (-C)$ a positive one) if $\tilde{\p}[\tilde{y}=+1|x] - \p[\tilde{y}_{q}=+1] > 0$.

Next we derive the probability of having $\tilde{\p}[\tilde{y}=+1|x] > \tilde{\p}[\tilde{y}_{q}=+1]$. When $n$ is sufficiently large, $\tilde{\p}[\tilde{y}_{q}=+1] \approx p_+ \cdot (1-e_+)+p_- \cdot e_{-} < \p[\tilde{y}=+1|x] = 1-e_+$. Define $Z_l$ as in Proof for Theorem \ref{thm:losscorrection}: the $l$ Bernoulli random variable that $Z_k = \mathds{1}(\tilde{y}(k) = +1), k \in [l]$. Then  using Hoeffding bound we prove that
\begin{align*}
    &\p\left[\tilde{\p}[\tilde{y}=+1|x] > p_+ \cdot (1-e_+)+p_- \cdot e_{-}\right]\\
    =&1-\p\left[\frac{\sum_{k \in [l]}Z_k}{l}  \leq p_+ \cdot (1-e_+)+p_- \cdot  e_-\right]\\
    \geq & 1-e^{ \frac{-2l}{\left(p_+ \cdot (1-e_+)+p_- \cdot  e_- - (1-e_+)\right)^2}}\\
    \geq&1-e^{ \frac{-2l}{p^2_{-} \cdot \left(1-e_+-e_-\right)^2}},
\end{align*}
completing the proof. Again the case with $y=-1$ is symmetric.
\end{proof}

\section*{Proof for Corollary \ref{cor:peerloss}}
\begin{proof}
Because we have shown that w.p. at least $1-e^{ \frac{-2l}{p^2_{sgn(-y)}\cdot \left(1-e_+-e_-\right)^2}}$, training with $\ell_{\text{PL}}$ on $x \in X_{S=l}$ returns $h(x)=y$, results a 0 individual excessive generalization error  $\err^{+}_l(\Pd,\A, x|S) = 0$. On the other hand, Theorem \ref{thm:noisy} informs us that $\err^{+}_l(\Pd,\A, x|S)$ is in the following order when memorizing the noisy labels:
\begin{align}
\Omega\left( \frac{l^2}{n^2} \textsf{weight}\left(\pi,\left[\frac{2}{3}\frac{l-1}{n-1}, \frac{4}{3}\frac{l}{n}\right] \right)\cdot \sum_{k \neq y} \tilde{\p}[\tilde{y}=k|x]\right) \label{eqn:cor2}
\end{align}
Taking the difference (Eqn.(\ref{eqn:cor2}) $- 0$) we proved the claim. 
\end{proof}

\section*{Proof for Theorem \ref{thm:peerloss:lower}}
\begin{proof}
Consider $y=+1$. 
As argued in the proof of Theorem \ref{thm:peerloss}, when $\tilde{\p}[\tilde{y}=+1|x] < \p[\tilde{y}_{q}=+1]$, we have $h(x)=-1$ returns a smaller loss. 
So the output $h(x)$ predicts $-1$, the wrong label. Now we show when $\tilde{\p}[\tilde{y}=+1|x] < \tilde{\p}[\tilde{y}_{q}=+1]$. Define $Z_l$ as in Proof for Theorem \ref{thm:losscorrection}: the $l$ Bernoulli random variable that $Z_k = \mathds{1}(\tilde{y}(k) = +1), k \in [l]$. Then 
\begin{align*}
&\p\left[\tilde{\p}[\tilde{y}=+1|x] < \tilde{\p}[\tilde{y}_{q}=+1]\right]\\
    =&\p\left[\tilde{\p}[\tilde{y}=+1|x] < p_+ (1-e_+)+p_- \cdot e_{-}\right]\\
    =&\p\left[\sum_{k \in [l]}1-Z_k  \geq l( p_+ \cdot e_+ +p_- \cdot (1-e_-))\right]\\
    =&  \p\left[\text{Bin}(l,e_+) \geq l\left( p_+ \cdot e_+ +p_- \cdot (1-e_-)\right)\right].
\end{align*}
Again using the tail bound for Bin, we prove that 
\begin{align*}
    & \p\left[\text{Bin}(l,e_+) \geq l( p_+ \cdot e_+ +p_- \cdot (1-e_-))\right] \\
    \geq & \frac{e^{-l \cdot D_{\text{KL}}(\frac{1}{2}\| e_+)}}{\sqrt{8l \left( p_+ \cdot e_+ +p_- \cdot (1-e_-)\right)\cdot \left(p_+ \cdot (1-e_+) +p_- \cdot e_-\right)}}.
\end{align*}
Note that $\left( p_+ \cdot e_+ +p_- \cdot (1-e_-)\right) + \left(p_+ \cdot (1-e_+) +p_- \cdot e_-\right) = 1$, and each term $\left( p_+ \cdot e_+ +p_- \cdot (1-e_-)\right)$ and $\left(p_+ \cdot (1-e_+) +p_- \cdot e_- \right)$ is positive. Therefore 
\[
\left( p_+ \cdot e_+ +p_- \cdot (1-e_-)\right) \cdot  \left(p_+ \cdot (1-e_+) +p_- \cdot e_-\right) \leq \frac{1}{4}.
\]
Therefore we conclude that 
\begin{align*}
    &\frac{e^{-l \cdot D_{\text{KL}}(\frac{1}{2}\| e_+)}}{\sqrt{8l \left( p_+ \cdot e_+ +p_- \cdot (1-e_-)\right)\cdot \left(p_+ \cdot (1-e_+) +p_- \cdot e_-\right)}}
    \geq \frac{1}{\sqrt{2l}}e^{-l \cdot D_{\text{KL}}(\frac{1}{2}\| e_+)},
\end{align*}
completing the proof.
\end{proof}

 \section{Figures}
 
 \section*{More examples for Figure \ref{fig:2d:ce}}
\begin{figure}[!ht]
\centering
\includegraphics[width=0.3\textwidth]{figures/bcewithlogits_clean.png}
\includegraphics[width=0.3\textwidth]{figures/bcewithlogits_random-0_2.png}
\includegraphics[width=0.3\textwidth]{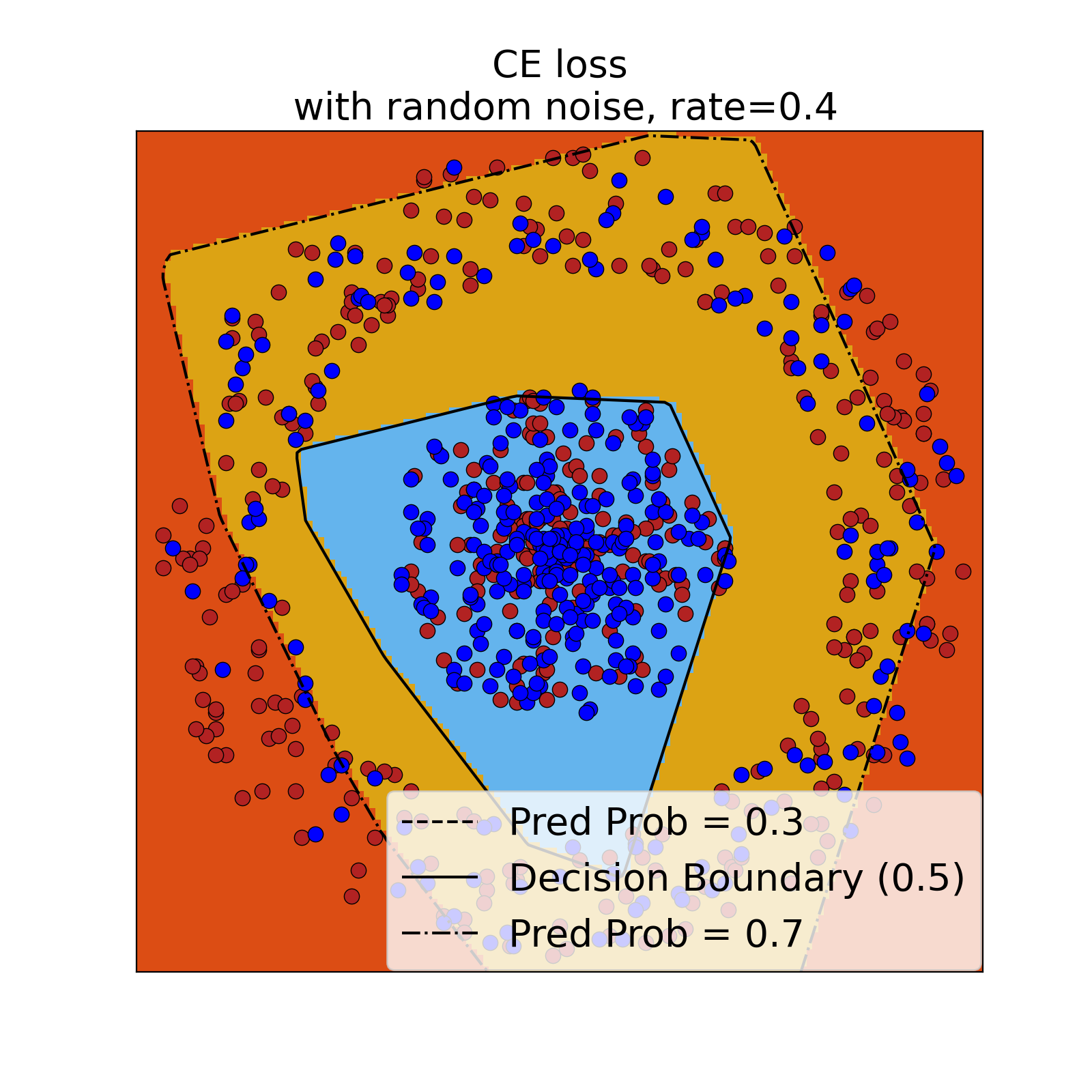}
\vspace{-0.2in}

\caption{A 2D example illustrating the memorization of noisy labels. Left panel: Training with clean labels. Middle panel: Training with random $20\%$ noisy labels. Right panel: Training with random $40\%$ noisy labels.}
\label{fig:2d:ce:full}
\end{figure}

 In Figure \ref{fig:2d:ce:full}, we provide one additional figure for training with 40\% random label noise for comparison.
 
 \section*{Details for generating Figure \ref{fig:cifar-memorization}}
 
The training uses ResNet-34 as the backbone with the following setups: mini-batch size (64), optimizer (SGD), initial learning rate (0.1), momentum (0.9), weight decay (0.0005), number of epochs (100) and learning rate decay (0.1 at 50 epochs). Standard data augmentation is applied to each dataset.
 
 The generation of instance-dependent label noise follows the steps in \citep{xia2020parts,cheng2020learning}:
 \begin{itemize}
    \item Define a noise rate (the global flipping rate) as $\varepsilon$.
    \item For each instance $x$, sample a $q$ from the truncated normal distribution $\mathbf{N}(\varepsilon, 0.1^{2}, [0, 1])$.
    \item  Sample parameters $W$ from the standard normal distribution. Compute $x \times W$.
     \item Then the final noise rate is a function of both $q$ and $x \cdot W$. 
 \end{itemize}

\section{Discussion}

\paragraph{Sample cleaning} Sample cleaning promotes the procedure of identifying possible wrong labels and removing them from training. If the identification is done correctly, the above procedure is effectively pushing $\tilde{\p}(\tilde{y})$ to the direction of the correct label, achieving a better generalization performance. From the other perspective, removing noisy labels helps the $h$ to de-memorize the noisy ones. 

The ability to identify the wrong labels has been more or less analyzed in the literature, but certainly would enjoy a more thorough and in-depth investigation. We view this as an important theoretical question for the community. Similar to the previous observations, we would like to caution the existence of rare examples with small $l$ - the previously introduced approaches have been shown to have a non-negligible chance of failing in such cases. We imagine this is probably true for sample cleaning, when the majority of a small number of noisy labels are in fact misleading. 
\paragraph{Understanding the difficulty of labeling
}
A better understanding of the possibility of handling noisy labels calls for immediate effort for understanding the probability of observing a wrong label for different instances. The salient challenge in doing so is again due to the missing of ground truth supervision information. One promising direction to explore is to leverage inference models \cite{liu2012variational} to infer the hidden difficulty factors in the generation of noisy labels.

\paragraph{Hybrid \& decoupled training}
Our results also point out a promising direction to treat samples from different regimes differently. While the existing approaches seem to be comfortable with the highly frequent instances (large $l$), the rare instances might deserve different handling. 

One thing we observed is rare samples suffer from a small $l$ and insufficient label information. While dropping rare samples is clearly hurting the generalization power, a better alternative would be to collecting multiple labels for these instances to boost the classifier's confidence in evaluating these instances.

\paragraph{Dataset effort} Last but not least, a concrete understanding of the effects of instance-dependent label noise would require a high-quality dataset that contains real human-level noise patterns. While there have been some recent efforts \cite{xiao2015learning,jiang2020beyond}, most of the evaluations and studies stay with the synthetic ones.


\end{document}